# Bayesian Structural Model Updating with Multimodal Variational Autoencoder

Tatsuya Itoi [a,*], Kazuho Amishiki [b], Sangwon Lee [a], Taro Yaoyama [a]

a School of Engineering, The University of Tokyo, 7-3-1 Hongo, Bunkyo-ku, Tokyo, 1138656, Japan

b School of Science, The University of Tokyo, 7-3-1 Hongo, Bunkyo-ku, Tokyo, 1130033, Japan

∗ Corresponding author.

E-mail address: itoi@g.ecc.u-tokyo.ac.jp (T. Itoi).

ABSTRACT

A novel framework for Bayesian structural model updating is presented in this study. The proposed method utilizes the surrogate unimodal encoders of a multimodal variational autoencoder (VAE). The method facilitates an approximation of the likelihood when dealing with a small number of observations. It is particularly suitable for high-dimensional correlated simultaneous observations applicable to various dynamic analysis models. The proposed approach was benchmarked using a numerical model of a single-story frame building with acceleration and dynamic strain measurements. Additionally, an example involving a Bayesian update of nonlinear model parameters for a three-degree-of-freedom lumped mass model demonstrates computational efficiency when compared to using the original VAE, while maintaining adequate accuracy for practical applications.

## 1. Introduction

The use of vibrational responses to monitor civil structures, including buildings, has been gaining attention in recent decades. To assess the condition and performance of structures, various methodologies have been proposed to update the parameters of dynamic analysis models using monitoring data. The dynamic analysis model that is originally based on assumptions made during the design phase requires adjustments to incorporate monitoring data. These adjustments are crucial for assessing the performance of structures during their lifetimes. Nevertheless, monitoring data, such as vibration responses, often lacks sufficient information on the parameters of dynamic analysis models [1], [2]. For instance, vibrational responses during minor earthquakes may not provide information regarding parameters related to nonlinear responses. Additionally, acceleration monitoring at floor level does not provide direct information on damage to structural or nonstructural members. The limited information from monitoring data introduces uncertainty into the estimation of the model parameters. Consequently, structural model-updating techniques should account for this lack of information and the associated uncertainties in the estimated model parameters. Such uncertainty can be described through the probability distribution of model parameters or by using a family of plausible model parameters that have been numerically estimated, instead of searching for the best estimate [3], [4], [5]. The performance of structures can be effectively assessed without bias [6], [7] by obtaining the probability distribution of the model parameters for a dynamic analysis model that incorporates these uncertainties. Thus, the Bayesian approach is effective for this purpose.

Within the realm of the Bayesian approach, modal-based approaches for Bayesian model updating have been developed [8], [9], [10], [11]. These methods utilize modal parameters, such as modal frequencies and mode shapes, which are estimated from the observed structural responses for model updating. Time-domain approaches [12], [13], [14], which are suitable for situations in which the structural properties vary over time, and frequency-domain approaches [15], [16], in which both the input and output need to be measured, have also been proposed. Furthermore, methods incorporating machine learning techniques within the Bayesian updating framework, such as the active learning of kriging [17], a pattern-recognition method [18], and the Bayesian inference approach [19], have been proposed. However, these methods frequently involve approximations and simplifications, primarily because of their theoretical limitations. This is particularly true when high-dimensional and correlated observations, such as time- or frequency-domain data, are applied to Bayesian structural model updating [20], [21]. To approximate the posterior distribution of the model parameters specifically in this context, an approximate Bayesian computation was developed [22], [23], [24]. The use of a variational autoencoder (VAE) for Bayesian structural model updating [20] is a promising approach to approximating Bayesian computations for a small number of observations, although it is computationally demanding.

In this study, a multimodal VAE [25], [26], [27] was employed to develop a versatile framework for Bayesian structural model updating. A multimodal VAE is an enhanced version of the original VAE [28] and is a notable tool in the field of machine learning. The proposed methodology provides a robust means of model updating for various structural models and measurements, effectively addressing uncertainties in the updated model parameters. A numerical example is presented as a benchmark for the proposed method, demonstrating the updating of model parameters of a single-story frame building structure equipped with acceleration and dynamic strain measurements. In the proposed approach, which utilizes a multimodal VAE, the computational demands are significantly lower than that of the original VAE, while the accuracy required for practical applications is maintained. An additional numerical example demonstrating the updating of the model parameters for a three-degree-of-freedom nonlinear structural model highlights the advantages and areas for further refinement.

## 2. Bayesian model updating with VAE

### 2.1 Overview of VAE and multimodal VAE
#### 2.1.1 Overview of VAE

The VAE, a generative model that represents both generative and inference models, was originally proposed by Kingma and Welling [28]. It is a prominent example of generative models, which are a type of unsupervised learning. The VAE is specifically designed to replicate the training data by learning an underlying latent representation that is not directly observable but is inferred from the observed data, thereby capturing the hidden structure within the dataset. Figure 1 is a graphical representation of the VAE, which facilitates the inference of latent representations and enables the generation of original data using neural networks.

Considering data $\mathbf{x}_k \in \mathbb{R}^M$ ($k = 1, \cdots, N_\mathbf{x}$), a row vector with $M$ entries, and the collection of various data simultaneously obtained at different $N_\mathbf{x}$ points, $\mathbf{X} = \left[\mathbf{x}_1^T \; \mathbf{x}_2^T \; \cdots \; \mathbf{x}_{N_\mathbf{x}}^T\right]^T \in \mathbb{R}^{N_\mathbf{x} \times M}$, where superscript T denotes the transposition, and $\mathbf{x}_k$ typically represents high dimensional, correlated, and simultaneously observed data such as time histories, and frequency response functions. When the VAE is trained in the space of the structural analysis model, the data $\mathbf{X}$ represent multiple outputs from a single-response analysis. In the VAE, each data $\mathbf{X}$ can be reconstructed or transformed from a $N_\mathbf{z}$ dimensional random variable $\mathbf{z} \in \mathbb{R}^{N_\mathbf{z}}$ via the generative model known as the decoder. The solid line in Fig. 1 shows graphical representation of the decoder. Variable $\mathbf{z}$ is referred to as latent variable. The dimensions of $\mathbf{z}$, $N_\mathbf{z}$, are usually smaller than the number of elements in $\mathbf{X}$, $N_\mathbf{x} \times M$. A sample of $\mathbf{X}$, denoted as $\mathbf{X}^{(k)}$ ($k = 1, \cdots, N$), is generated following the probability distribution $p(\mathbf{X}|\mathbf{z})$, using a sample of $\mathbf{z}$, denoted as $\mathbf{z}^{(k)}$ ($k = 1, \cdots, N$), which follows the probability distribution $p(\mathbf{z})$.

Subsequently, the inverse of the generative model, that is, the posterior distribution of $\mathbf{z}$, $p(\mathbf{z}|\mathbf{X})$, is expressed as follows:

$$p(\mathbf{z}|\mathbf{X}) = \frac{p(\mathbf{X}|\mathbf{z})p(\mathbf{z})}{\int p(\mathbf{X}|\mathbf{z})p(\mathbf{z})d\mathbf{z}} \qquad (1)$$

This corresponds to the inference of the latent variable $\mathbf{z}$ hidden behind the data $\mathbf{X}$. The dashed line in Fig. 1 denotes the graphical model representation of the inference model $q_\Phi(\mathbf{z}|\mathbf{X})$ known as the encoder which is the approximation of the posterior distribution of $\mathbf{z}$, i.e., $p(\mathbf{z}|\mathbf{X})$. The decoder and encoder of the VAE are modeled using neural networks. $\Phi$ denotes the parameters related to the neural network of the encoder.

The training of the VAE aims to better represent the training dataset with minimal loss of information by minimizing the Kullback-Leibler distance between the probability distributions of a random variable $\mathbf{z}$ and the standard normal distribution and by simultaneously minimizing the error to reconstruct $\mathbf{X}$ from $\mathbf{z}$. The objective function $\mathcal{L}(\mathbf{X})$ for training the VAE includes two components: the reconstruction term and the Kullback-Leibler distance term, which has the opposite sign of the loss function:

$$\mathcal{L}(\mathbf{X}) = \mathbb{E}_{q_\Phi(\mathbf{z}|\mathbf{X})}[\log p(\mathbf{X}|\mathbf{z})] - D_{\mathrm{KL}}[q_\Phi(\mathbf{z}|\mathbf{X}) \| p(\mathbf{z})] \qquad (2)$$

where $\mathbb{E}_{q_\Phi(\mathbf{z}|\mathbf{X})}[\blacksquare]$ is the expectation operator with respect to $q_\Phi(\mathbf{z}|\mathbf{X})$, and $D_{\mathrm{KL}}[q_\Phi(\mathbf{z}|\mathbf{X}) \| p(\mathbf{z})]$ is the Kullback-Leibler distance of $q_\Phi(\mathbf{z}|\mathbf{X})$ from $p(\mathbf{z})$. The first and second terms on the right side of Eq. (2) correspond to the reconstruction term of the decoder and the regularization term, respectively. This minimization ensures that the probability distribution of $\mathbf{z}$, where each element is independent of the others, closely approximates the standard normal distribution. Hereafter, $p(\mathbf{z})$ is approximated as the probability density function of a normal distribution with mean $\mathbf{0}$ and covariance matrices $\mathbf{I}$, $N(\mathbf{0}, \mathbf{I})$.

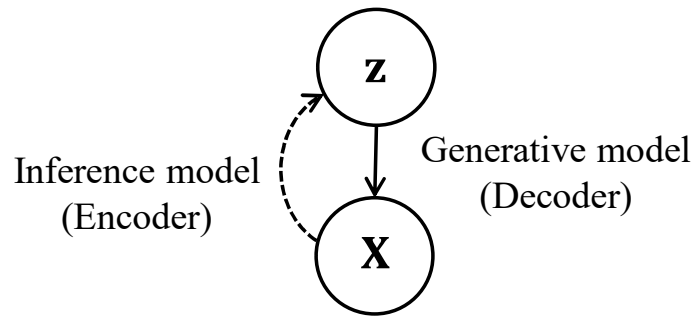

Fig. 1 Graphical model representation of VAE

2.1.2 Overview of multimodal VAE

Multimodal VAE is an enhanced version of the original VAE. The joint multimodal VAE (JMVAE) is a multimodal VAE classified as a joint VAE [25]. In the JMVAE-kl proposed by Suzuki et al. [26], [27], a joint VAE capable of learning from multiple modalities with different dimensions was realized by introducing surrogate unimodal encoders, each of which is a VAE for a single modality, in addition to a joint model. In the context of machine learning, modalities refer to different ways of representing or expressing the subject matter. The use of the JMVAE-kl allows for bidirectional generation between modalities. Consider that JMVAE-kl is trained in the space of dynamic analysis model $\mathcal{M}$, as shown in Fig. 2, using the results of the response analyses. Then, the model parameters $\boldsymbol{\theta}$ and their corresponding responses $\mathbf{X}$ are regarded as multiple modalities that can be generated from each other through a shared representation of latent variables $\mathbf{z}$. The arrows in the encoders shown in Fig. 2 (a) represent the inference process from the model parameters $\boldsymbol{\theta}$ and their corresponding responses $\mathbf{X}$ to the latent variable $\mathbf{z}$. The arrows in the decoder shown in Fig. 2 (b) represent the generating process from the latent variable $\mathbf{z}$ to both the model parameters $\boldsymbol{\theta}$ and their responses $\mathbf{X}$. The decoder and encoders of the multimodal VAE are designed using neural networks, similar to the structure of the original VAE.

The dashed lines in Fig. 2 (a) indicate that learning was conducted to minimize the Kullback-Leibler distance between the probability distributions of the outputs from the encoders of the unimodal and joint models, leading to nearly identical outputs **z**. The objective function for training the JMVAE-kl is as follows:

$$\mathcal{L}_2(\boldsymbol{\theta}, \mathbf{X}) = \mathcal{L}_1(\boldsymbol{\theta}, \mathbf{X}) - D_{\mathrm{KL}}[q_{\Phi_{\boldsymbol{\theta}}}(\mathbf{z}|\boldsymbol{\theta}) \| p(\mathbf{z}|\boldsymbol{\theta}, \mathbf{X})] - D_{\mathrm{KL}}[q_{\Phi_{\mathbf{X}}}(\mathbf{z}|\mathbf{X}) \| p(\mathbf{z}|\boldsymbol{\theta}, \mathbf{X})]. \tag{3}$$

$\mathcal{L}_1(\boldsymbol{\theta}, \mathbf{X})$ in the right-hand side of Eq. (3) can be expressed in a manner similar to Eq. (2), as follows:

$$\mathcal{L}_1(\boldsymbol{\theta}, \mathbf{X}) = \mathbb{E}_{q_{\Phi}(\mathbf{z}|\boldsymbol{\theta}, \mathbf{X})}[\log p(\boldsymbol{\theta}, \mathbf{X}|\mathbf{z})] - D_{\mathrm{KL}}[q_{\Phi}(\mathbf{z}|\boldsymbol{\theta}, \mathbf{X}) \| p(\mathbf{z})] \tag{4}$$

where the first and second terms on the right side of Eq. (4) correspond to the reconstruction term of the decoder (Fig. 2 (b)), and the regularization term, respectively.

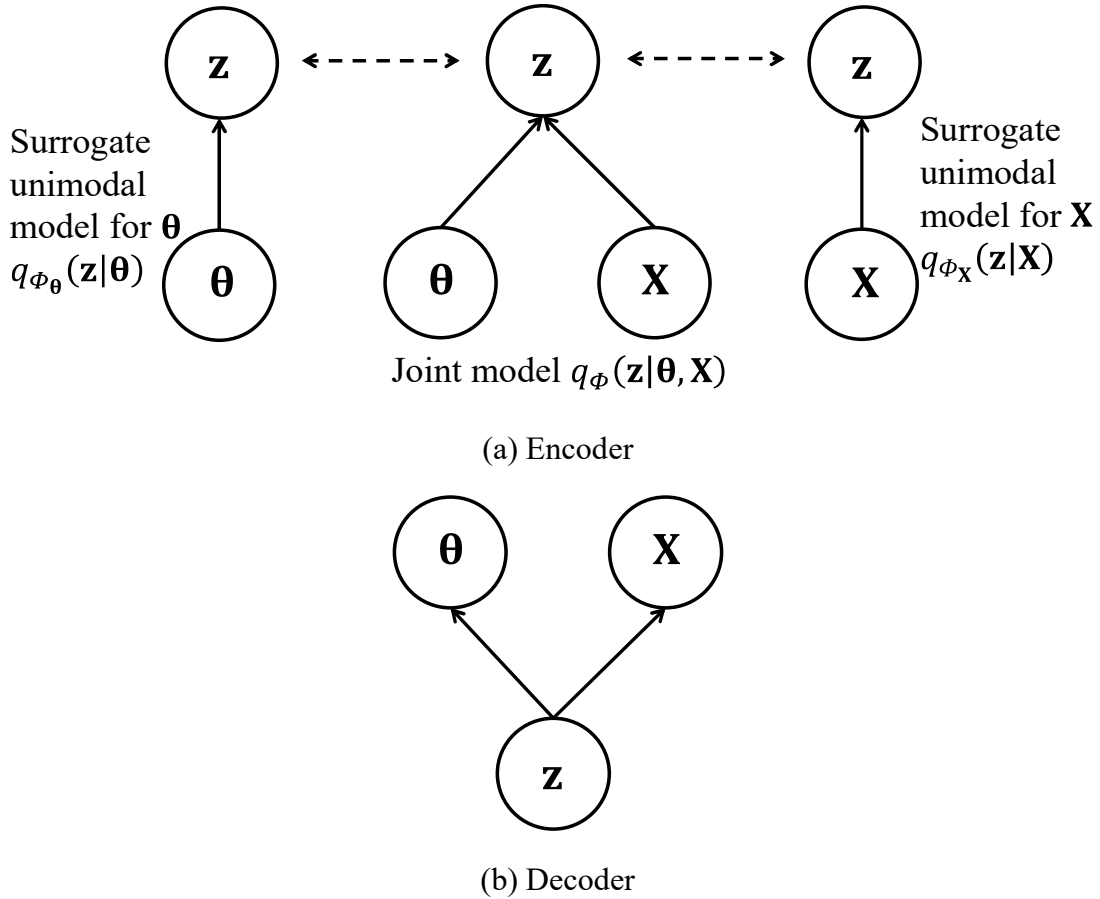

(a) Encoder

(b) Decoder

Fig. 2 Graphical model representation of JMVAE-kl

2.2 Likelihood estimation using multimodal VAE

Using Bayes' theorem, the updated probability distribution of the model parameters $\boldsymbol{\theta}$ based on the observation $\mathbf{X}_{obs}$, $p(\boldsymbol{\theta}|\mathbf{X}_{obs})$, is obtained as follows [5]:

$$p(\boldsymbol{\theta}|\mathbf{X}_{obs}) = c_1 p(\mathbf{X}_{obs}|\boldsymbol{\theta}) p(\boldsymbol{\theta}), \qquad (5)$$

where, $p(\boldsymbol{\theta})$ is the prior probability distribution of the model parameters $\boldsymbol{\theta}$. $p(\mathbf{X}_{obs}|\boldsymbol{\theta})$ is the probability of obtaining the observed data $\mathbf{X}_{obs}$ given the model parameters $\boldsymbol{\theta}$, known as the likelihood function. $c_1 = 1/p(\mathbf{X}_{obs})$ is the normalizing constant.

By introducing the latent variable $\mathbf{z}$ of the VAE [20], the likelihood $p(\mathbf{X}_{obs}|\boldsymbol{\theta})$, as seen on the right side of Eq. (5), can be transformed as follows:

$$p(\mathbf{X}_{obs}|\boldsymbol{\theta}) = \int p(\mathbf{X}_{obs}|\mathbf{z}) p(\mathbf{z}|\boldsymbol{\theta}) d\mathbf{z}, \qquad (6)$$

where the dimension of $\mathbf{z}$ is smaller than the number of elements in $\mathbf{X}_{obs}$. The utilization of the latent variable $\mathbf{z}$ of the VAE, as shown in Eq. (6), by reconstructing data from a latent variable with minimal loss of information, minimizes the loss of information in the observed data $\mathbf{X}_{obs}$. It is also important to note that the components of the latent variable $\mathbf{z}$ are mutually independent standard normal variables. Accordingly, they are simpler to handle than typical correlated components of $\mathbf{X}_{obs}$.

By applying Bayes' theorem, the term $p(\mathbf{X}_{obs}|\mathbf{z})$ on the right-hand side of Eq. (6) can be expressed as:

$$p(\mathbf{X}_{obs}|\mathbf{z}) = \frac{p(\mathbf{z}|\mathbf{X}_{obs}) p(\mathbf{X}_{obs})}{p(\mathbf{z})} = c_2 \cdot \frac{p(\mathbf{z}|\mathbf{X}_{obs})}{p(\mathbf{z})} \qquad (7)$$

where $c_2$ is a normalizing constant. Substituting Eq. (7) into Eq. (6) yields

$$p(\mathbf{X}_{obs}|\boldsymbol{\theta}) = c_2 \int_{-\infty}^{\infty} \frac{p(\mathbf{z}|\mathbf{X}_{obs}) p(\mathbf{z}|\boldsymbol{\theta})}{p(\mathbf{z})} d\mathbf{z}, \qquad (8)$$

where $p(\mathbf{z}|\mathbf{X}_{obs})$ in Eq. (8) is approximated by the output from the encoder of the VAE using $\mathbf{X}_{obs}$ as the input. As described in Eq. (1), $p(\mathbf{z})$ is approximated to be the probability density function of a normal distribution with a mean $\mathbf{0}$ and covariance $\mathbf{I}$, $N(\mathbf{0}, \mathbf{I})$. $p(\mathbf{z}|\boldsymbol{\theta})$ is the conditional probability density function given the model parameters $\boldsymbol{\theta}$. To calculate $p(\mathbf{X}_{obs}|\boldsymbol{\theta})$, according to Eq. (8), the Markov chain Monte Carlo (MCMC) sampling is adopted. Regarding the use of the original VAE (refer to Fig. 1) to obtain $p(\mathbf{z}|\boldsymbol{\theta})$, each sample of $\mathbf{X}$ corresponding to each sample of model parameters $\boldsymbol{\theta}$ is obtained using response analysis. Then, $p(\mathbf{z}|\boldsymbol{\theta})$ is derived by feeding each sample of $\mathbf{X}$ into the encoder of the VAE. This process, which involves response analysis, requires relatively large computational resources to calculate the likelihood, particularly in the case of the seismic response analysis of structures with high degrees of freedom.

In this study, we propose the use of multimodal VAE for the approximation of likelihood $p(\mathbf{X}_{obs}|\boldsymbol{\theta})$ as an alternative to the original VAE [20]. As introduced in Subsection 2.1, the use of surrogate unimodal encoders for $\boldsymbol{\theta}$ and $\mathbf{X}$ in JMVAE-kl, which is trained in the space of the response analyses, enables approximation of $p(\mathbf{z}|\mathbf{X}_{obs})$ and $p(\mathbf{z}|\boldsymbol{\theta})$. The approximation for $p(\mathbf{z}|\mathbf{X}_{obs})$ in Eq. (8) was obtained using the surrogate unimodal encoder of $\mathbf{X}$, $q_{\Phi_X}(\mathbf{z}|\mathbf{X})$ by substituting $\mathbf{X}_{obs}$ into $q_{\Phi_X}(\mathbf{z}|\mathbf{X})$. The approximation of $p(\mathbf{z}|\boldsymbol{\theta})$ in Eq. (8) was obtained using the surrogate unimodal encoder of $\boldsymbol{\theta}$, $q_{\Phi_{\boldsymbol{\theta}}}(\mathbf{z}|\boldsymbol{\theta})$. Notably, when training the VAE, the dataset must be compiled such that the parameters $\boldsymbol{\theta}$ of the sample response analysis models are uniformly distributed. Computation time is required to create training data, because seismic response analyses must be performed using sample response analysis models. Likelihood calculation does not require significant additional time when a multimodal VAE is used, which is one advantage of the approach. This is because employing the surrogate unimodal encoder for $\boldsymbol{\theta}$, $q_{\Phi_{\boldsymbol{\theta}}}(\mathbf{z}|\boldsymbol{\theta})$, to approximate $p(\mathbf{z}|\boldsymbol{\theta})$ eliminates the need for seismic response analyses for each MCMC sample, a process that is computationally demanding when the original VAE is used. However, employing the surrogate unimodal encoder for $\boldsymbol{\theta}$ may affect accuracy, which will be explored in Section 4.

## 3. Fundamental benchmark for the proposed method

### 3.1 Problem description

The performance of the proposed method for estimating the likelihood was benchmarked using the procedure shown in Fig. 3. In this benchmark problem, earthquake monitoring was performed in a ground-truth building by observing both the input ground motion, $\mathbf{y}_{obs}$ and the associated responses, $\mathbf{X}_{obs}$. This process is illustrated in the box labeled '1. Ground-truth structure (model parameter $\boldsymbol{\theta}_{gt}$)' in Fig. 3 (explained in more detail in Subsection 3.2). Then, a dataset of responses $\mathbf{X}$ corresponding to the model parameters $\boldsymbol{\theta}$ for training the JMVAE-kl was compiled through a series of response analyses, utilizing the observed input ground motion $\mathbf{y}_{obs}$ as an input. This process is illustrated in the section labeled '2. The space of the response analysis model $\mathcal{M}$' in Fig. 3 (explained in more detail in Section 3.3). Finally, the family of plausible model parameters $\boldsymbol{\theta}$ that effectively reproduces the observed response $\mathbf{X}_{obs}$ was identified by estimating the likelihood $p(\mathbf{X}_{obs}|\boldsymbol{\theta})$ using Eq. (8) and the posterior distribution $p(\boldsymbol{\theta}|\mathbf{X}_{obs})$ using Eq. (5), as illustrated at the bottom right of Fig. 3. To calculate the likelihood $p(\mathbf{X}_{obs}|\boldsymbol{\theta})$, surrogate unimodal encoders both for $\mathbf{X}$ and $\boldsymbol{\theta}$ of JMVAE-kl, $q_{\Phi_X}(\mathbf{z}|\mathbf{X})$ and $q_{\Phi_{\boldsymbol{\theta}}}(\mathbf{z}|\boldsymbol{\theta})$, were used, as discussed in Section 2 (explained in more detail in Subsection 3.4).

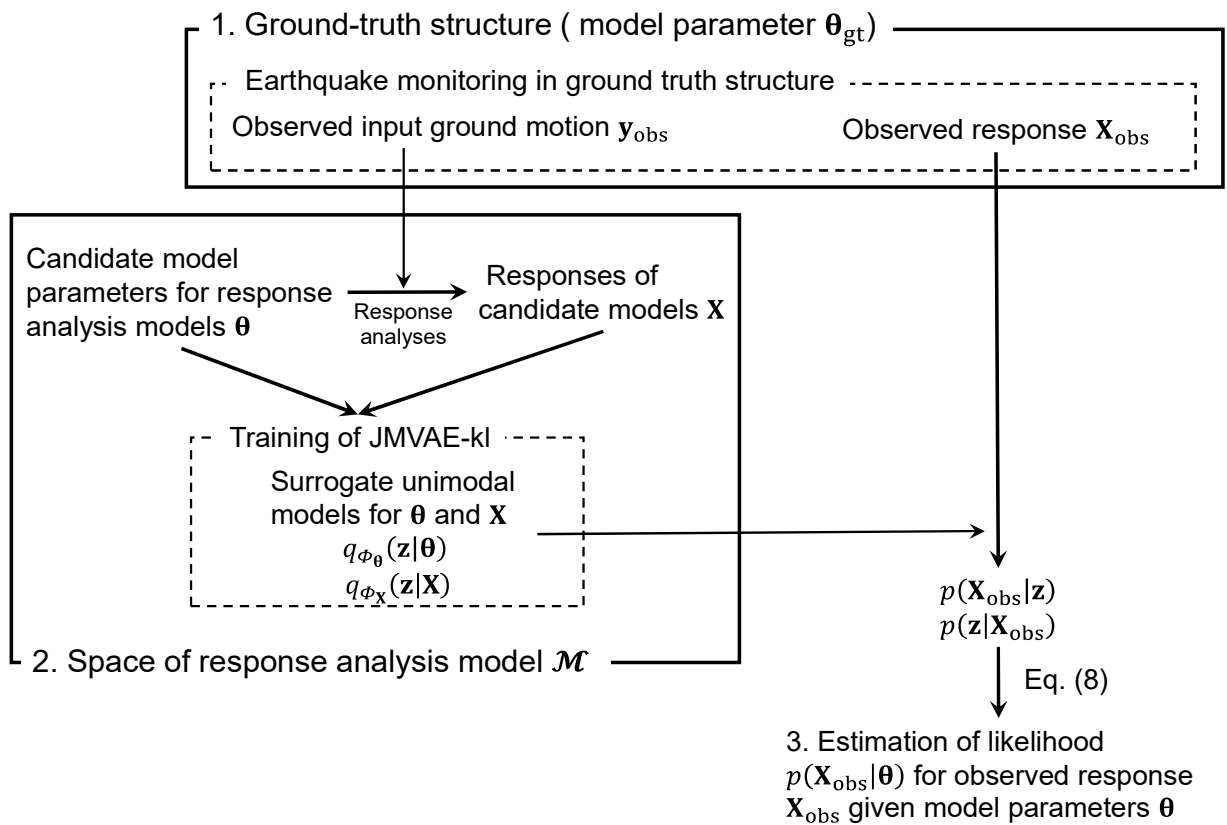

Fig. 3 Framework of proposed structural model updating procedure for fundamental benchmark problem

3.2 Model description

The simplified frame model shown in Fig. 4 was used as the response analysis model. The model is a plane-moment-resisting frame model with rotational springs at the bottom of the columns. The two columns, each with upper and lower part with different sections, were connected to the roof deck by a beam. The cross-sectional area, $A$; the second moment of area, $I$, of all the elements; and the section modulus, $Z$, of the column members are assumed, as shown in Fig. 4. The response analysis was conducted on OpenSees [29], an open-source software for structural analysis, using its wrapper for Python, OpenSeesPy [30]. The model was based on the real-scale, single-span, single-story building structure presented by Yaoyama et al. [31]. Addressing Bayesian model updating using actual experimental data obtained from a real-scale structure is a more challenging issue than that presented in this study. This aspect should be explored in future studies. When applying the proposed method to actual structures and observed data, the selection of an appropriate model (i.e., model selection) becomes an additional issue.

The ground-truth model parameters $\boldsymbol{\theta}_{gt}$ for the four different cases are listed in Table 1. The variation in the rotational springs represents the damage to the anchor bolts connecting the base plate to the foundation. The mass of the frame structure and roof deck was assumed to be 7000 kg, which was allocated to Nodes 3 and 4. The damping ratio was 0.02 for the first mode and was proportional to the stiffness. In this example, the rotational stiffness of each rotational spring was assumed to be unknown. The effect of the roof deck on the beam was modeled as an increase in the bending stiffness and was assumed to be unknown. The equivalent bending stiffness, $(EI)_{eq}$, considering the effect of the roof deck, was modeled as follows:

$$(EI)_{eq} = R \cdot (EI)_{b}, \qquad (9)$$

where $(EI)_{b}$ represents the bending stiffness of the beam, and $R$ is a constant that models the increase in bending stiffness owing to the deck. For simplicity, the effect of the roof deck on the beam axial stiffness was not considered. The other parameters related to the dynamic analysis model are assumed to be known.

In this real-scale structure [31], acceleration measurements were conducted at the basement and roof, whereas dynamic strain measurements were performed on all the columns. The model assumed similar measurement conditions: acceleration measurement at Node 4, dynamic strain measurements at the bottom of Elements 1 and 2, and measurements at the top of Elements 3 and 4. Dynamic strain $\varepsilon_l(t)$ ($l = 1, \cdots, 4$) at location $l$ typically includes the effects of the static stress owing to the constant value of the self-weight, in addition to those of the dynamic seismic response. When only the effects of the dynamic seismic response were considered, the dynamic strain $\varepsilon_l(t)$ was calculated from the results of response analysis as follows:

$$\varepsilon_l(t) = \frac{M_l(t)}{EZ}, \qquad (10)$$

where $M_l(t)$ represents the time history of the bending moment at each location, $E$ is Young's modulus of the steel (205 GPa), and $Z_l$ is the elastic section modulus of each column.

For the input ground motion, the El Centro acceleration time history record from the 1940 Imperial Valley earthquake (north-south component) was used, with a sampling frequency of 50 Hz and a duration of 40.96 seconds, as shown in Fig. 5. The acceleration time histories at Node 4, along with the dynamic-strain time histories at the bottom of Elements 1 and 2 and the top of Elements 3 and 4, were stored as the responses, as shown in Fig. 6. Gaussian noise with an S/N ratio of 40 dB was added to both the input acceleration and response time histories. The acceleration and dynamic strains exhibited similar temporal characteristics, and the amplitudes of the dynamic strains varied between the observation points depending on the conditions of rotational stiffness. The Fourier amplitude spectral ratios of the response to the input acceleration were obtained, as shown in Fig. 7. Hereafter, the logarithm of these Fourier amplitude spectral ratios from 0.12 to 12.6 Hz, comprising 512 data points, were used as $\mathbf{X}_{obs}$. The size of $\mathbf{X}_{obs}$ was 5×512. The design of the measurement and types of data used for the model updating $\mathbf{X}_{obs}$ were variable, which affected the posterior distribution of the model parameters $\boldsymbol{\theta}$, $p(\boldsymbol{\theta}|\mathbf{X}_{obs})$. The optimization of these choices warrants further investigation.

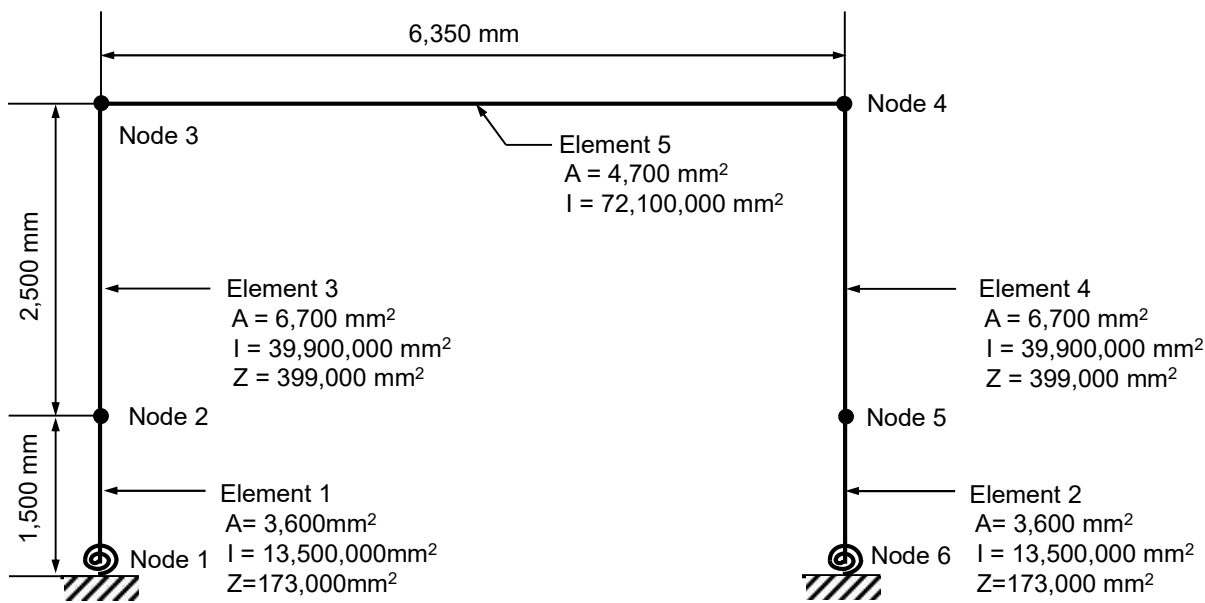

Fig. 4 Ground-truth model of frame structure

Table 1 Parameters of ground-truth models $\boldsymbol{\theta}_{gt}$

|  | Ratio of $(EI)_{eq}$ to $(EI)_b$ | Stiffness of rotational spring at Node 1 [kNm/rad] | Stiffness of rotational spring at Node 6 [kNm/rad] |
|---|---|---|---|
| Case 1 | 2.0 | 1000 | 1000 |
| Case 2 | 2.0 | 1000 | 5000 |
| Case 3 | 2.0 | 5000 | 1000 |
| Case 4 | 2.0 | 5000 | 5000 |

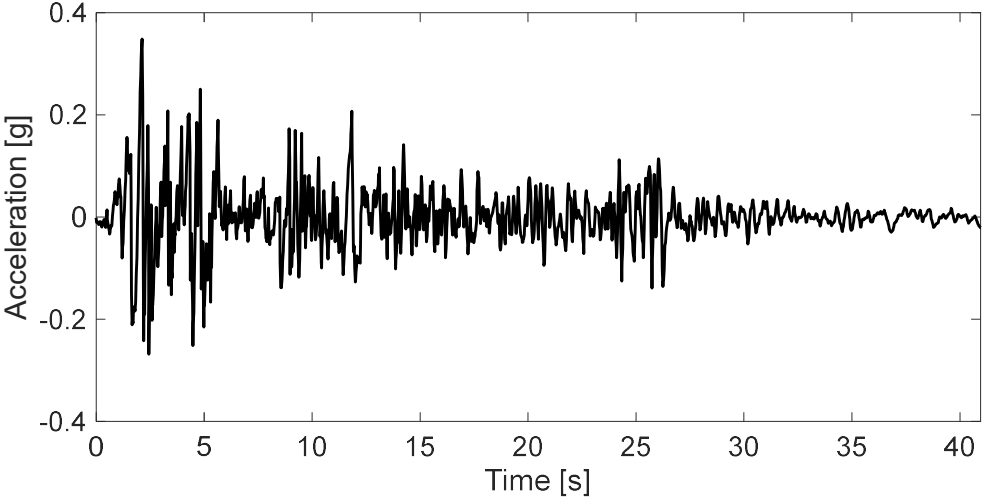

Fig. 5 Acceleration time history of input ground motion

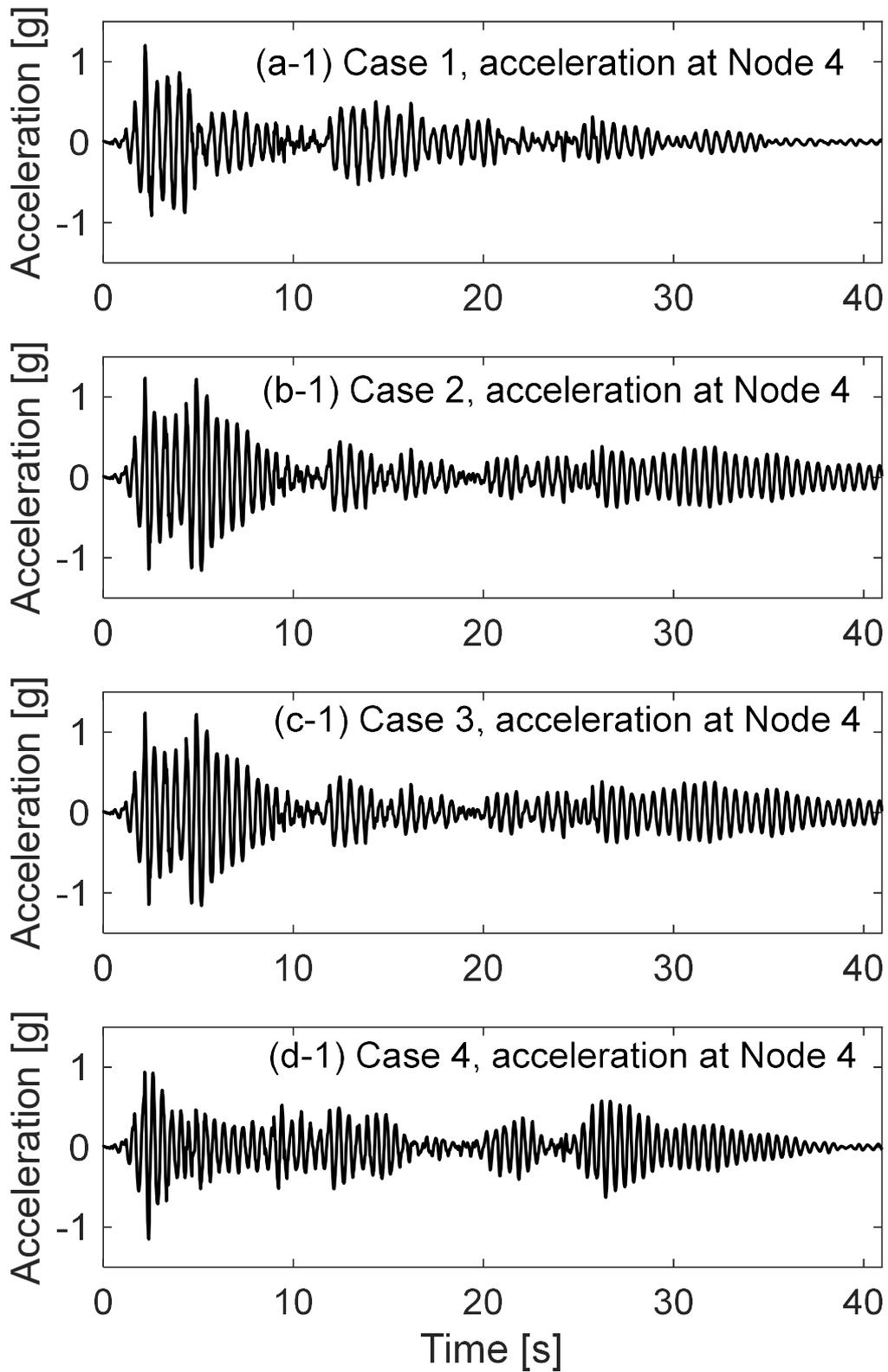

Fig. 6 Time histories of responses in ground-truth models

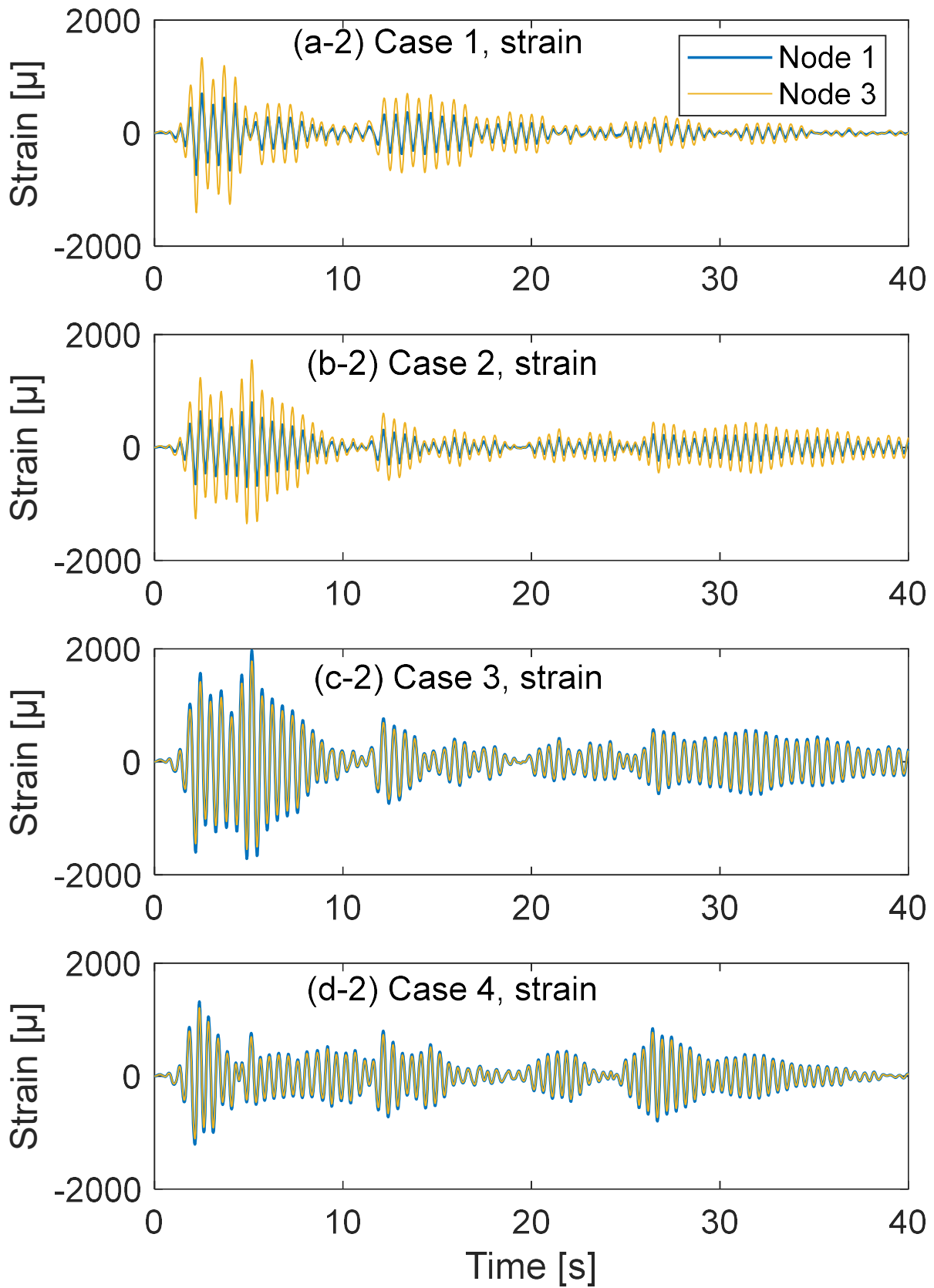

Fig. 6 Time histories of responses in ground-truth models (cont.)

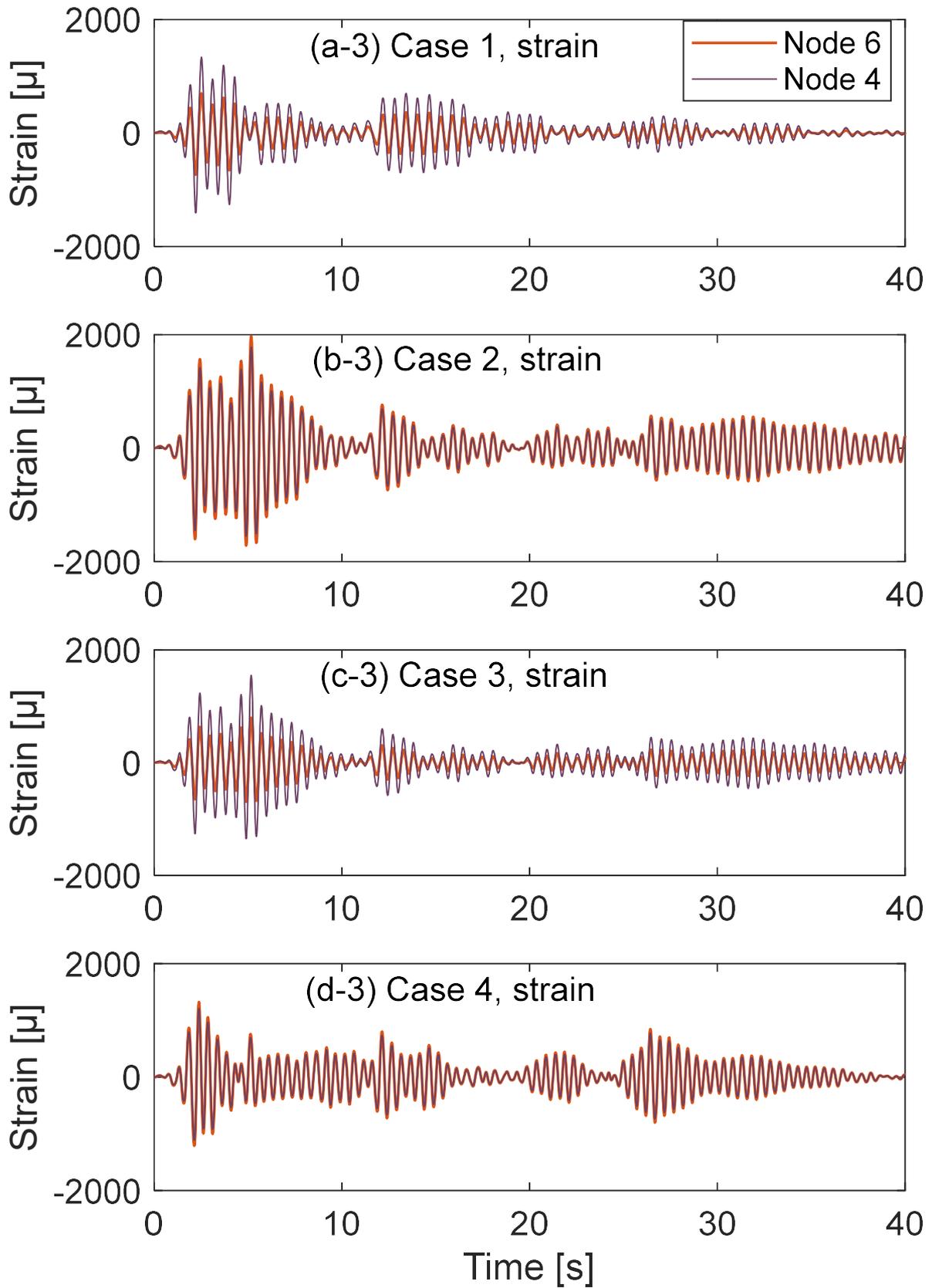

Fig. 6 Time histories of responses in ground-truth models (cont.)

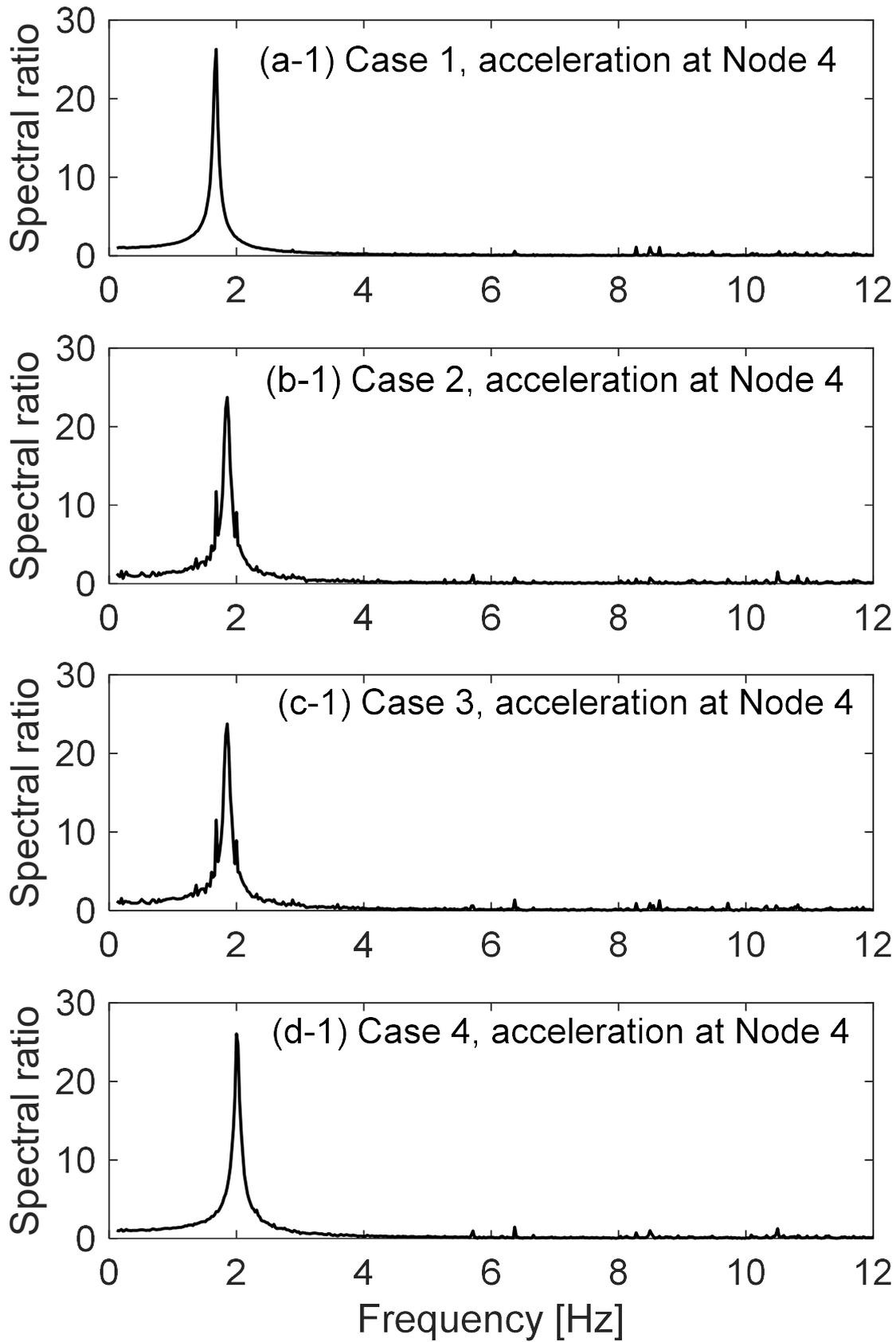

Fig. 7 Fourier amplitude spectral ratios of response to input acceleration for ground-truth models

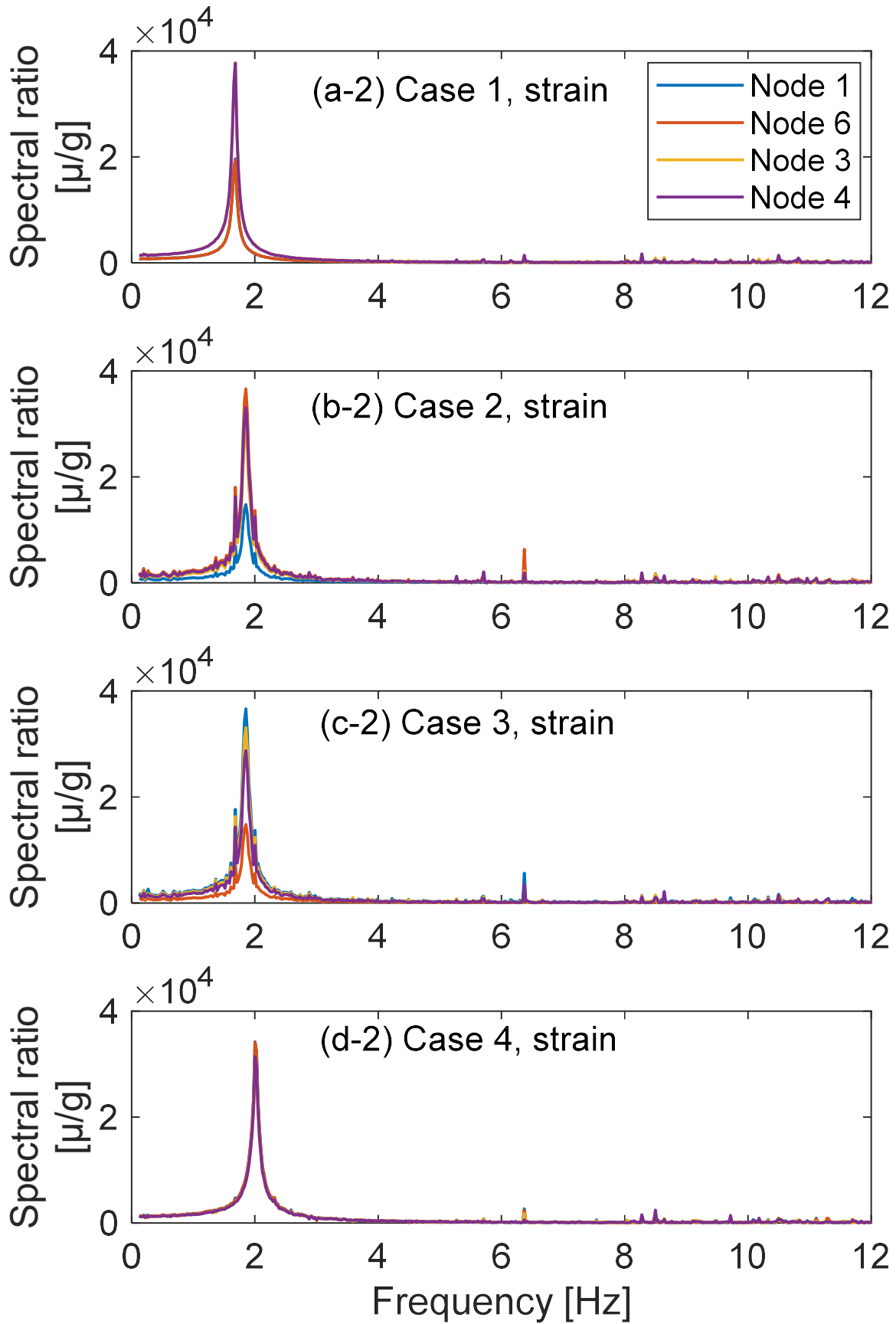

Fig. 7 Fourier amplitude spectral ratios of response to input acceleration for ground-truth models (cont.)

3.3 Training VAE in the space of response analysis model

Response analyses were conducted for various model parameters of the frame structure to construct a dataset to train the VAE. A total of $10^4$ pseudo-random numbers were generated for the model parameter samples based on the conditions listed in Table 2. For the input ground motion, the El Centro acceleration time history record from the 1940 Imperial Valley earthquake (north-south component) was used, which was identical to that observed in the ground-truth model described in Subsection 3.2. The acceleration time histories with Gaussian noise and an S/N ratio of 40 dB stored as responses were identical to the ground-truth model observations. The logarithm of the Fourier amplitude spectral ratios of the responses from 0.1 to 12.6 Hz (512 data points) was used as the training dataset.

Diagrams of the encoder, surrogate unimodal encoders, and decoder of the JMVAE-kl are shown in Figs. 8, 9, and 10, respectively. Fig. 8 illustrates the architecture of the encoder, which corresponds to the joint model $q_\Phi(\mathbf{z}|\mathbf{\theta},\mathbf{X})$ shown in Fig. 2(a). It took both $\mathbf{\theta}$ (a vector with three entries) and $\mathbf{X}$ (a matrix with a size of 5×512) as inputs. After passing $\mathbf{X}$ through four convolutional layers with a leaky rectified linear unit (ReLU) activation followed by three linear layers with a leaky ReLU activation, and $\mathbf{\theta}$ through three linear layers with a leaky ReLU activation, the outputs for $\mathbf{\theta}$ and $\mathbf{X}$ were merged. Then, the four linear layers with a leaky ReLU activation (excluding the last layer), output mean and variance of $\mathbf{z}$, $\mathbf{\mu_z}$ and $\mathbf{\sigma_z}^2$ (vectors with ten entries). Fig. 9 (a) and (b) present the structure of the surrogate unimodal encoders for $\mathbf{\theta}$ and $\mathbf{X}$, corresponding to surrogate unimodal models, $q_{\Phi_\theta}(\mathbf{z}|\mathbf{\theta})$ and $q_{\Phi_X}(\mathbf{z}|\mathbf{X})$, in Fig. 2(a). For the surrogate unimodal encoder for $\mathbf{\theta}$, seven linear layers were employed to output the mean and variance of $\mathbf{z}$, $\mathbf{\mu_z}$ and $\mathbf{\sigma_z}^2$. For the surrogate unimodal encoder of $\mathbf{X}$, four convolutional layers and six linear layers were used. Leaky ReLU activation was also used as the joint model.

Fig. 10 presents the architecture of the decoder, which received a sample of $\mathbf{z}$ as input and reconstructed $\mathbf{\theta}$ and $\mathbf{X}$ using neural networks. Leaky ReLU activation was also utilized. The decoder received $\mathbf{z}$ (a vector with ten entries) as the input, followed by four linear layers. The sequence was then divided into two parts: one consisted of three linear layers to reconstruct the mean and variance of $\mathbf{\theta}$ (a vector with six entries, denoted as $\mathbf{\theta}'$ in Fig. 10) and the other consisted of three linear layers followed by four convolutional layers to reconstruct the mean and variance of $\mathbf{X}$ (a matrix with a size of 10×512 and denoted as $\mathbf{X}'$ in Fig. 10).

The training dataset comprised 40,000 samples. Regarding the number of dimensions of the latent variable $\mathbf{z}$, the reconstruction error stabilized when the dimension of $\mathbf{z}$ exceeded eight. An increase in the dimension of $\mathbf{z}$ increased the reconstruction performance while increasing the computational costs. Accordingly, the number of dimensions of the latent variable $\mathbf{z}$ was set to 10. The VAE model was trained to minimize the loss function using an Adaptive Moment Estimation (Adam) optimizer [34]. The Adam optimizer is suitable for problems that are complex in terms of the amount of data and/or parameters; it has been widely adopted in deep learning applications. It dynamically adjusts the individual learning rates for different parameters based on the estimates of the first and second moments of the gradient. The batch size, initial training rate, and epoch size were set to 64, 0.0001, and 1000, respectively. The training parameters of the VAE are listed in Table 3.

Table 2 Range and probability distribution of parameters for the training dataset used for frame model

|  | Ratio of $(EI)_{eq}$ to $(EI)_b$ | Stiffness of rotational spring at Node 1 [kNm/rad] | Stiffness of rotational spring at Node 6 [kNm/rad] |
|---|---|---|---|
| Lower bound | 1 | 10 | 10 |
| Upper bound | 10 | $10^4$ | $10^4$ |
| Probability distribution | Uniform distribution in logarithmic axis | Uniform distribution in logarithmic axis | Uniform distribution in logarithmic axis |

Table 3 Training parameters for VAE used for frame model

| Batch size | 64 |
|---|---|
| Training rate | 0.0001 |
| Epoch size | 1000 |

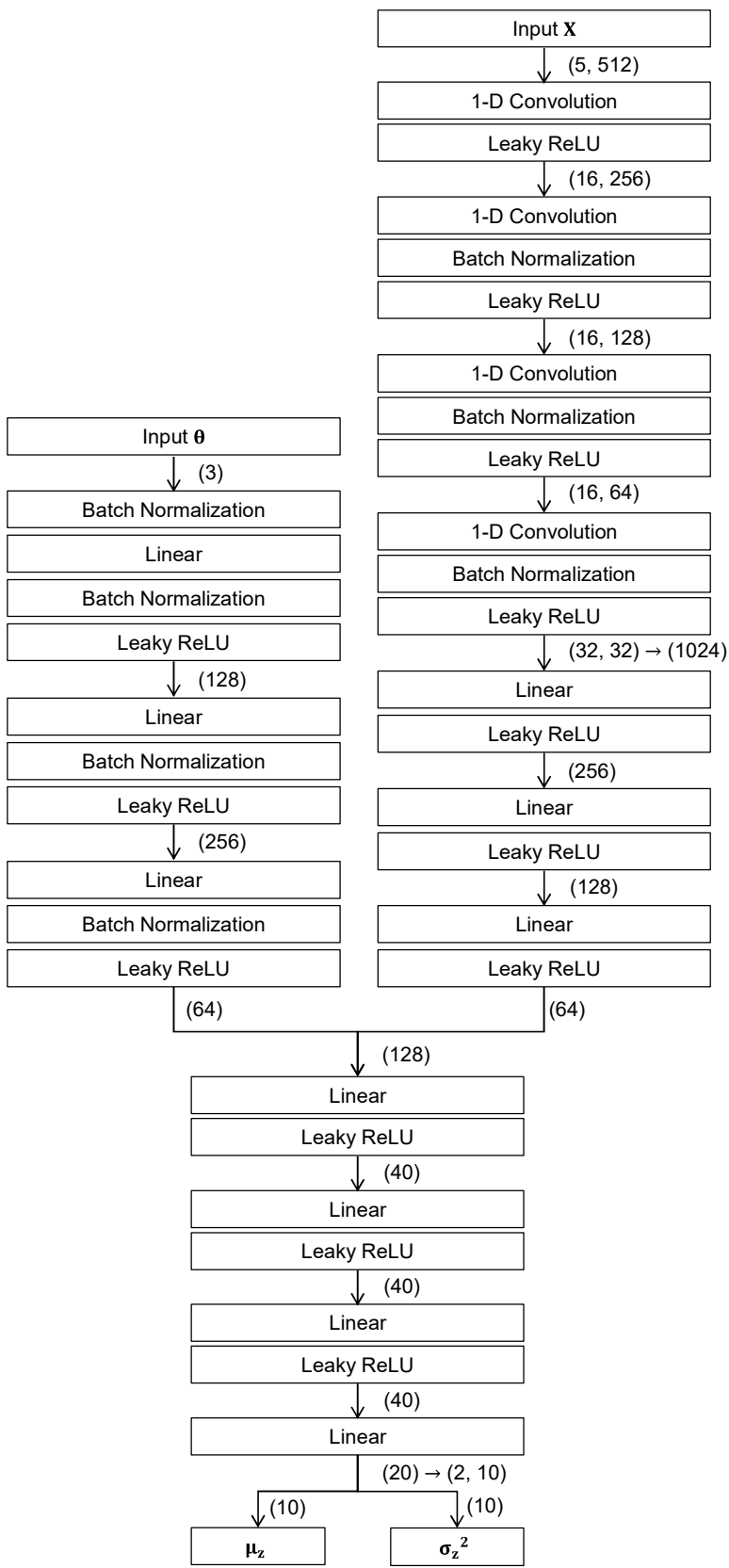

Fig. 8 Diagram of the encoder of JMVAE-kl used for frame model

(a) Surrogate unimodal model for θ    (b) Surrogate unimodal model for **X**

Fig. 9 Diagram of the surrogate unimodal encoders of JMVAE-kl used for frame model

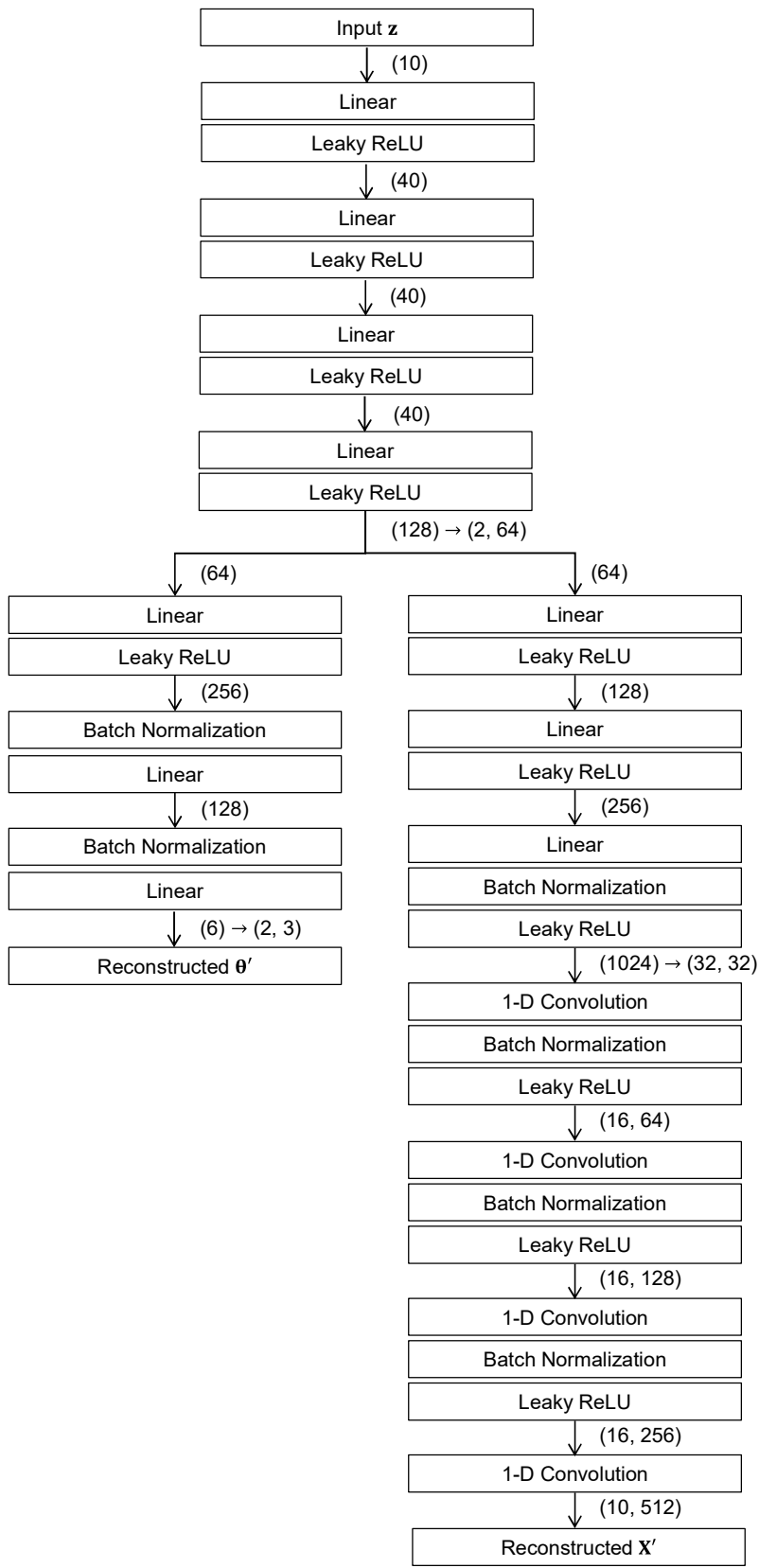

Fig. 10 Diagram of decoder of JMVAE-kl used for frame model

3.4 Posterior distribution of the model parameters

The sampling from the posterior distribution of the model parameters, $p(\boldsymbol{\theta}|\mathbf{X}_{\text{obs}})$, as expressed in Eq. (5), was performed using the MCMC method with the Metropolis-Hastings algorithm [32], [33]. In this study, a uniform distribution was assumed for the prior distribution $p(\boldsymbol{\theta})$ for simplification. In the MCMC process, the normal distribution was used as the proposal distribution, and the standard deviation was set to 0.1 times the parameter range. The first 10000 samples were discarded as the burn-in samples. The samples were thinned by retaining every 100th sample. A total of 1000 samples of the posterior distribution for $\boldsymbol{\theta}$ were obtained.

The posterior cumulative distribution function of the updated model parameters $\boldsymbol{\theta}$ obtained using the samples from MCMC are shown in Fig. 11. The estimated model parameters are close to the ground truth values in all four cases. The uncertainty in the estimated model parameters is also presented, with a wider range of $(EI)_{\text{eq}}/(EI)_{\text{b}}$ values than those of the rotational stiffness. The difference arose because information on the rotational stiffness could be obtained from the strains at the top and bottom of the columns, whereas information on $(EI)_{\text{eq}}/(EI)_{\text{b}}$ was obtained by combining information on the estimated rotational stiffnesses and observed strains. Therefore, the accuracy of $(EI)_{\text{eq}}/(EI)_{\text{b}}$ depended on the estimated rotational stiffness. Consequently, the uncertainty in $(EI)_{\text{eq}}/(EI)_{\text{b}}$ was more significant than that in the rotational stiffness.

The proposed Bayesian updating method successfully quantified uncertainty in the estimated model parameters. The degree of uncertainty depended on whether the observed data include information about the parameters, which, in turn, could depend on the arrangement and types of measurements, number of measurement points, and characteristics of the observed data. As discussed in Subsection 3.2, the design and optimization of the measurement and types of data $\mathbf{X}_{\text{obs}}$ affect the uncertainty of the estimated model parameters, and optimizing these choices warrants further discussion. The ability to evaluate these uncertainties is the strength of this method.

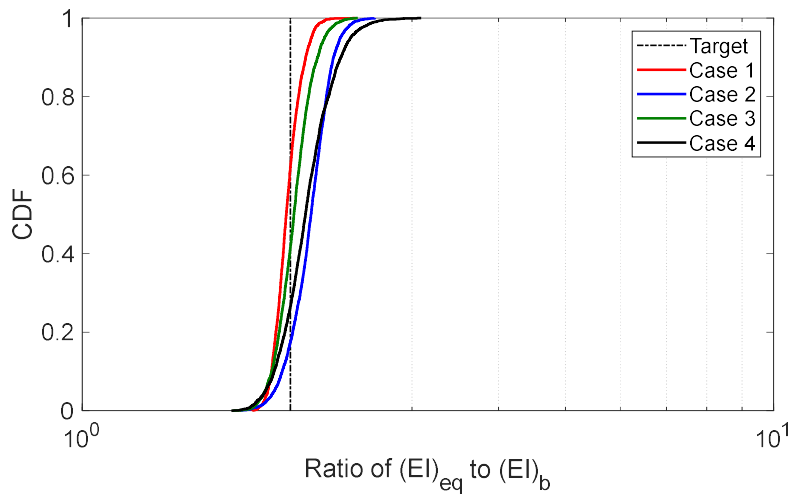

(a) Ratio of $(EI)_{eq}$ to $(EI)_b$

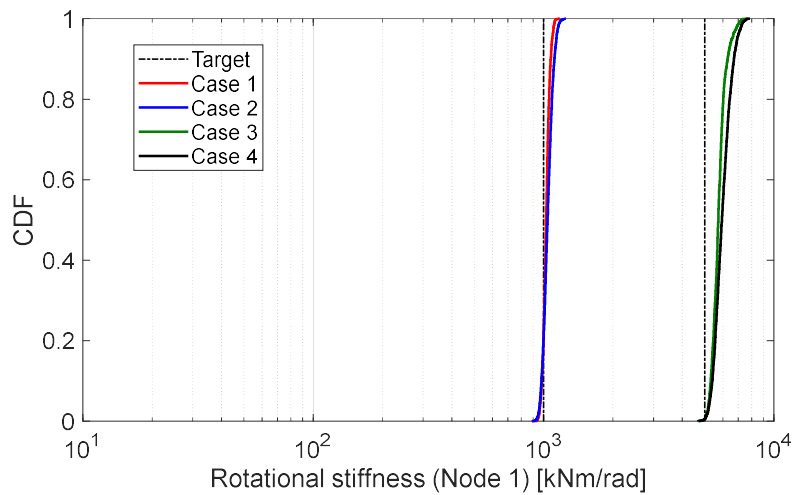

(b) Rotational stiffness of rotation spring at Node 1

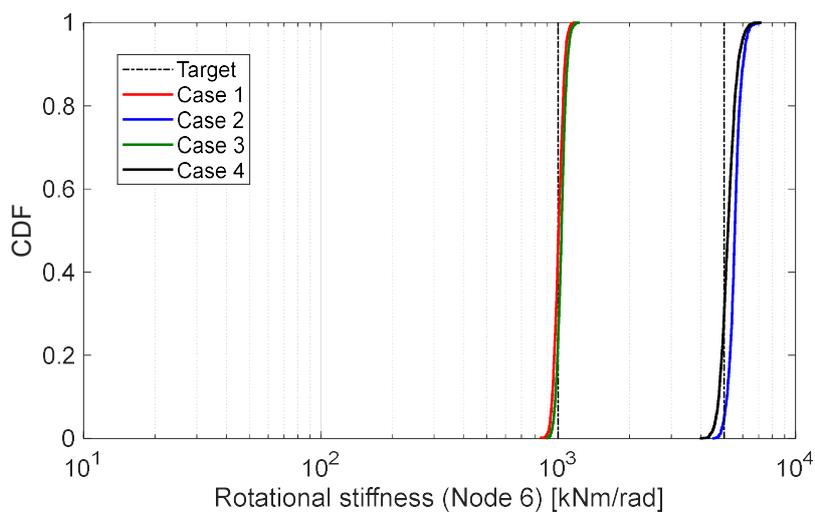

(c) Rotational stiffness of rotation spring at Node 6

Fig. 11 Cumulative distribution function of the posterior distribution of model parameters

## 4. Comparison between multimodal VAE and original VAE

### 4.1 Model description

In this section, the accuracy and computational demands of the proposed method, which uses a multimodal VAE, are compared with those of the original VAE approach [20]. The comparison was based on a three-degree-of-freedom lumped mass model subjected to ground motion, as shown in Fig. 12. A three-story reinforced concrete structure was assumed and acceleration measurements were conducted on each floor. The restoring force characteristics were modeled using a modified Takeda slip model [35]. The model parameters included the initial stiffness of the first, second, and third story, $k_i$ ($i = 1,2,3$), crack displacement $d_c$, yield displacement $d_y$, post-crack stiffness of each story $\alpha_c k_i$ ($i = 1,2,3$), post-yield stiffness of each story $\alpha_y k_i$ ($i = 1,2,3$), an index $\gamma$ to determine the unloading stiffness $k_d$, and an index $\lambda$ to define slip stiffness $k_s$. The parameters $d_c$, $d_y$, $\alpha_c$, $\alpha_y$, $\gamma$ and $\lambda$ were assumed to be uniform across all stories, whereas the initial stiffnesses $k_i$ ($i = 1,2,3$) were different across stories. Stiffnesses $k_d$ and $k_s$ were defined as follows:

$$k_d = \frac{Q_{\max}}{d_y - d_c} \cdot \left|\frac{d_{\max}}{d_y}\right|^{-\gamma} \quad (11)$$

$$k_s = \frac{Q_y - Q_c}{d_{\max} - d_0} \cdot \left|\frac{d_{\max}}{d_y}\right|^{-\lambda} \quad (12)$$

where $d_{\max}$ and $Q_{\max}$ are the maximum displacement and the corresponding restoring force, respectively. The model parameters $d_c$, $d_y$, $\alpha_c$, $\alpha_y$, $\gamma$, and $\lambda$ for the ground-truth model were assumed, as shown in Table 3. The masses of the first, second, and third stories $m_i$ ($i = 1, 2, 3$) were assumed to be known and set to be $7.5 \times 10^4$, $7.3 \times 10^4$, and $5.5 \times 10^4$ kg, respectively. For simplification, the damping ratio was assumed to be known and set to be 0.04 for the first mode, which was assumed to be proportional to the instantaneous stiffness.

Table 3 Parameters of modified Takeda slip model assumed for ground-truth model

| $k_1$ [kN/mm] | $k_2$ [kN/mm] | $k_3$ [kN/mm] | $d_c$ [mm] | $d_y$ [mm] | $\alpha_c$ | $\alpha_y$ | $\gamma$ | $\lambda$ |
|---|---|---|---|---|---|---|---|---|
| 140 | 110 | 60 | 8 | 40 | 0.1 | 0.02 | 0.4 | 0.5 |

The performance of the proposed method at likelihood estimation was evaluated using the procedure shown in Fig. 3, as demonstrated in the example in Section 3. The input ground motion for the ground-truth model was the mainshock ground motion of the 2016 Kumamoto earthquake in Japan recorded at the KMMH16 station of KiK-net [36] in the east-west direction, as shown in Fig. 13. The sampling frequency was 100 Hz, and the duration was 100s. Response analysis was performed, and the responses obtained were treated as hypothetically observed data. To introduce observational noise, random numbers that followed a normal distribution with a mean of 0 and a standard deviation of 0.001 m/s$^2$ were added to the input and response time histories. Fig. 14 shows examples of the response time history of the ground-truth model, where Fig. 14 (a) illustrates the acceleration on the third floor and Fig. 14 (b) illustrates the interstory drift on the first floor. The drift of the first floor exceeded the crack displacement $d_c$ and yield displacement $d_y$ for a short period.

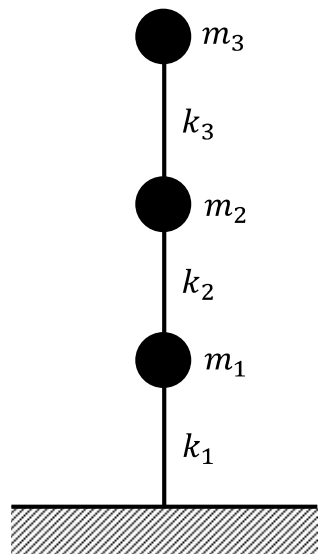

Fig. 12 Three-degree-of-freedom lumped mass model

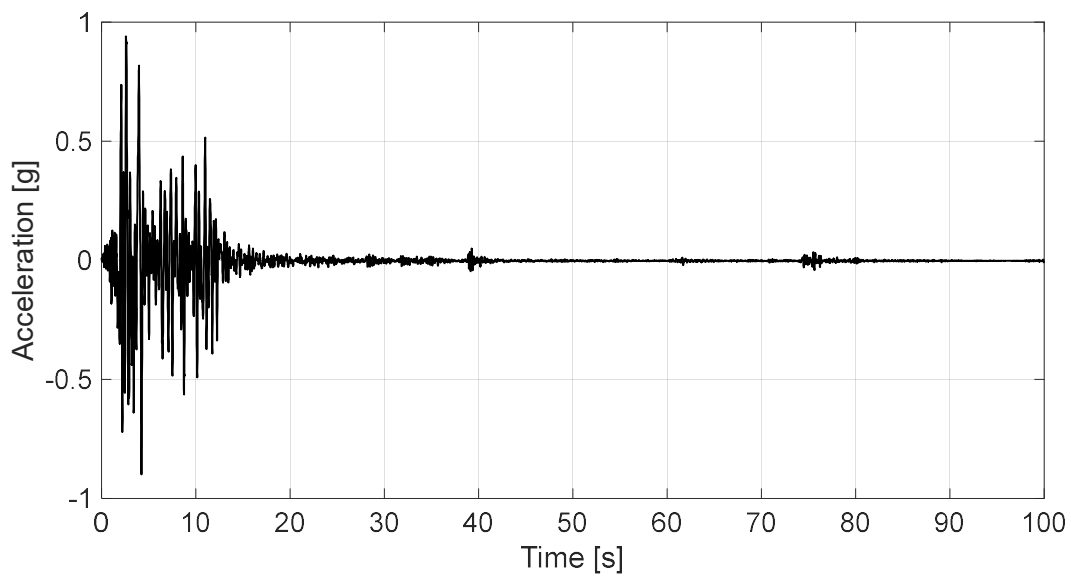

Fig. 13 Acceleration time history of input ground motion (East-West direction at KiK-net KMMH16 for the 2016 Kumamoto Earthquake)

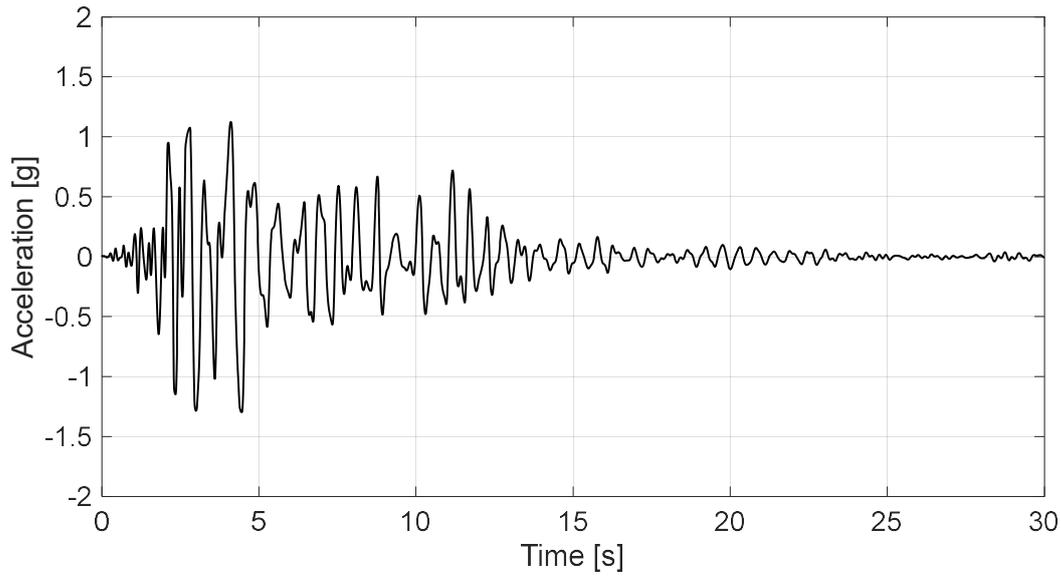

(a) Acceleration on third floor

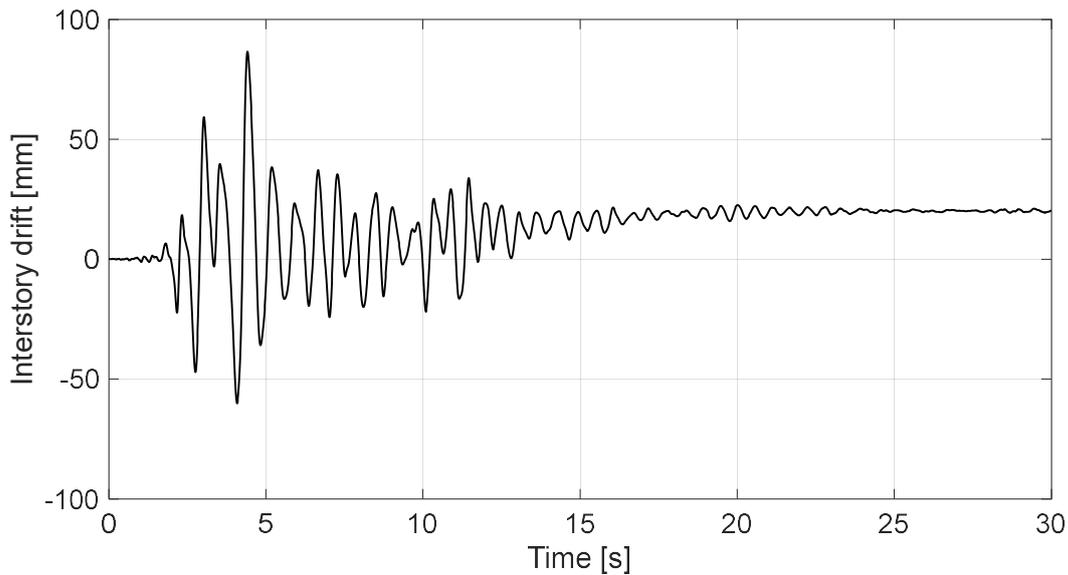

(b) Interstory drift on first floor

Fig. 14 Response time history of the ground-truth model

4.2 Training of VAE

A training dataset consisting of $10^5$ response analyses was created from $10^5$ different analysis models. The acceleration at the basement floor of the ground truth model discussed in Subsection 4.1 served as the input motion for the training dataset. The parameters of the response analysis models used for the training dataset were generated as uniformly distributed random numbers within the ranges listed in Table 4. For each floor, the frequency response function of the acceleration response to the input acceleration was obtained, comprising of the real and imaginary components within the range of 0.1 to 5.22 Hz (i.e., $2 \times 512$ points).

Imaginary components were also included to capture the nonlinear characteristics of the seismic responses. The frequency response functions were obtained for all three stories; therefore, the training dataset was structured into a matrix with dimensions $6 \times 512$.

Table 4 Ranges of model parameters used in creating the training data used for lumped mass model

|  | $k_1$ [kN/mm] | $k_2$ [kN/mm] | $k_3$ [kN/mm] | $d_c$ [mm] | $d_y$ [mm] | $\alpha_c$ | $\alpha_2$ | $\gamma$ | $\lambda$ |
|---|---|---|---|---|---|---|---|---|---|
| Upper bound | 200 | 160 | 120 | 20 | 80 | 0.25 | 0.05 | 1 | 1 |
| Lower bound | 100 | 60 | 20 | 2.5 | 20 | 0.05 | 0 | 0 | 0 |

The architecture of the multimodal VAE was designed as shown in Figs. 15–18. The architecture was designed as described in Section 3 with the addition of a residual block [37] to improve training performance. The joint encoder model (refer to Fig. 15) takes both $\boldsymbol{\theta}$ (a vector with nine entries) and $\mathbf{X}$ (a matrix with a size of $6 \times 512$) as inputs. Meanwhile, two surrogate unimodal encoders model (refer to Fig. 16) separately took $\boldsymbol{\theta}$ and $\mathbf{X}$ as their respective input. The decoder (refer to Fig. 17) has a symmetrical structure to the joint encoder, extending the input dimension of a residual block and fully connected (FC) layers, taking a sample of $\mathbf{z}$ (a vector with ten entries) as input to reconstruct $\boldsymbol{\theta}$ and $\mathbf{X}$.

The encoders had a structure that reduced the dimensions of $\mathbf{X}$ through residual blocks (Fig. 18(a)), constituting a convolutional neural network (CNN) layer and FC layers. The encoding residual blocks, which expanded the number of channels while reducing the data length via downsampling, are depicted in Fig. 18(a). The decoding residual blocks, which reduce the number of channels and increase the length of the data via upsampling, are shown in Fig. 18(b). This decision was based on the assumption that the degree of freedom of the variability of the data was equivalent to the number of parameters. The multimodal VAE was trained to minimize the loss function using the Adam optimizer with a learning rate of 0.0001. We set the batch size to 64 and stopped training after 1,000 epochs while confirming the decrease in the VAE loss function to ensure effective learning.

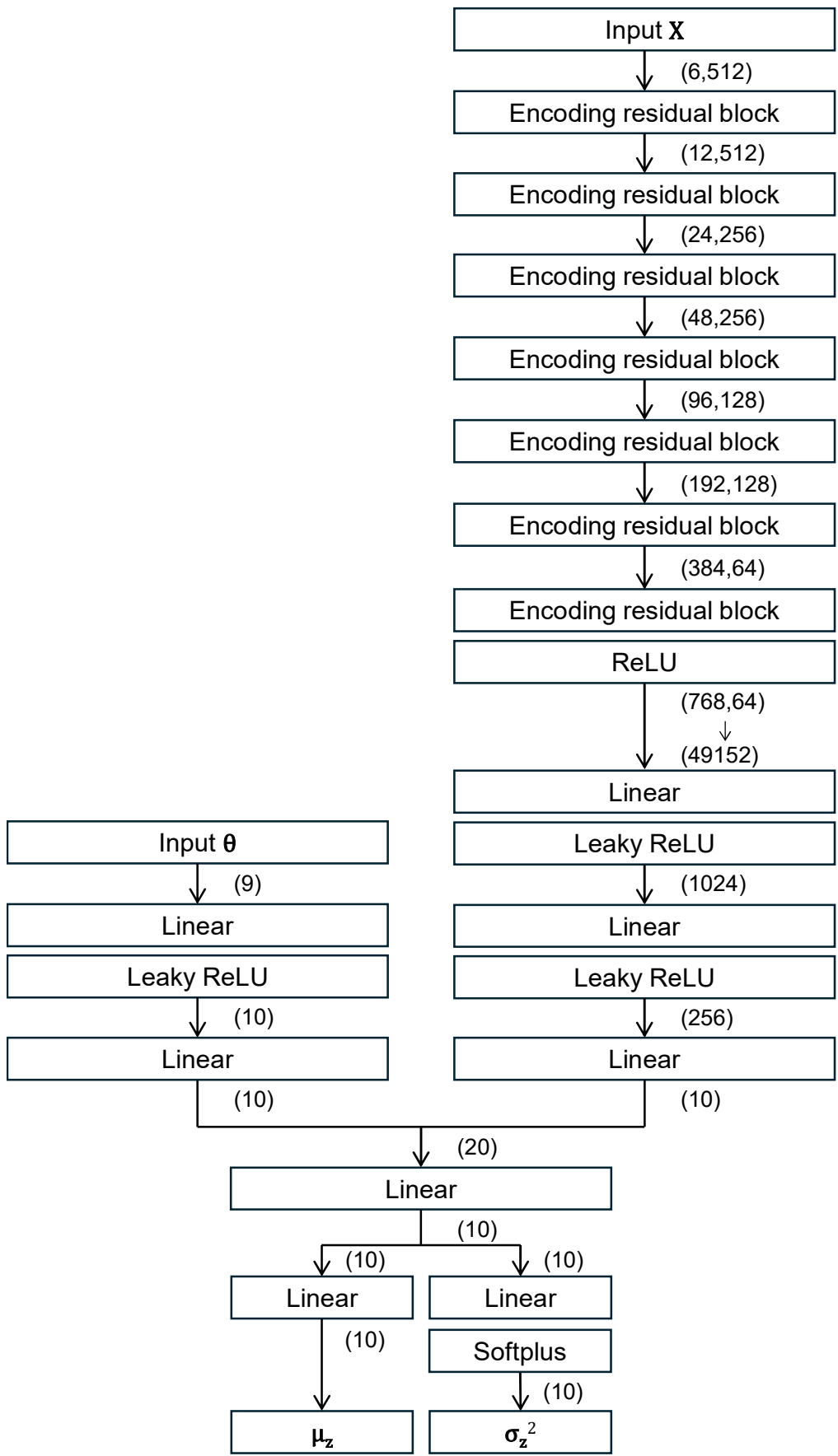

Fig. 15 Diagram of the encoder of JMVAE-kl used for lumped mass model

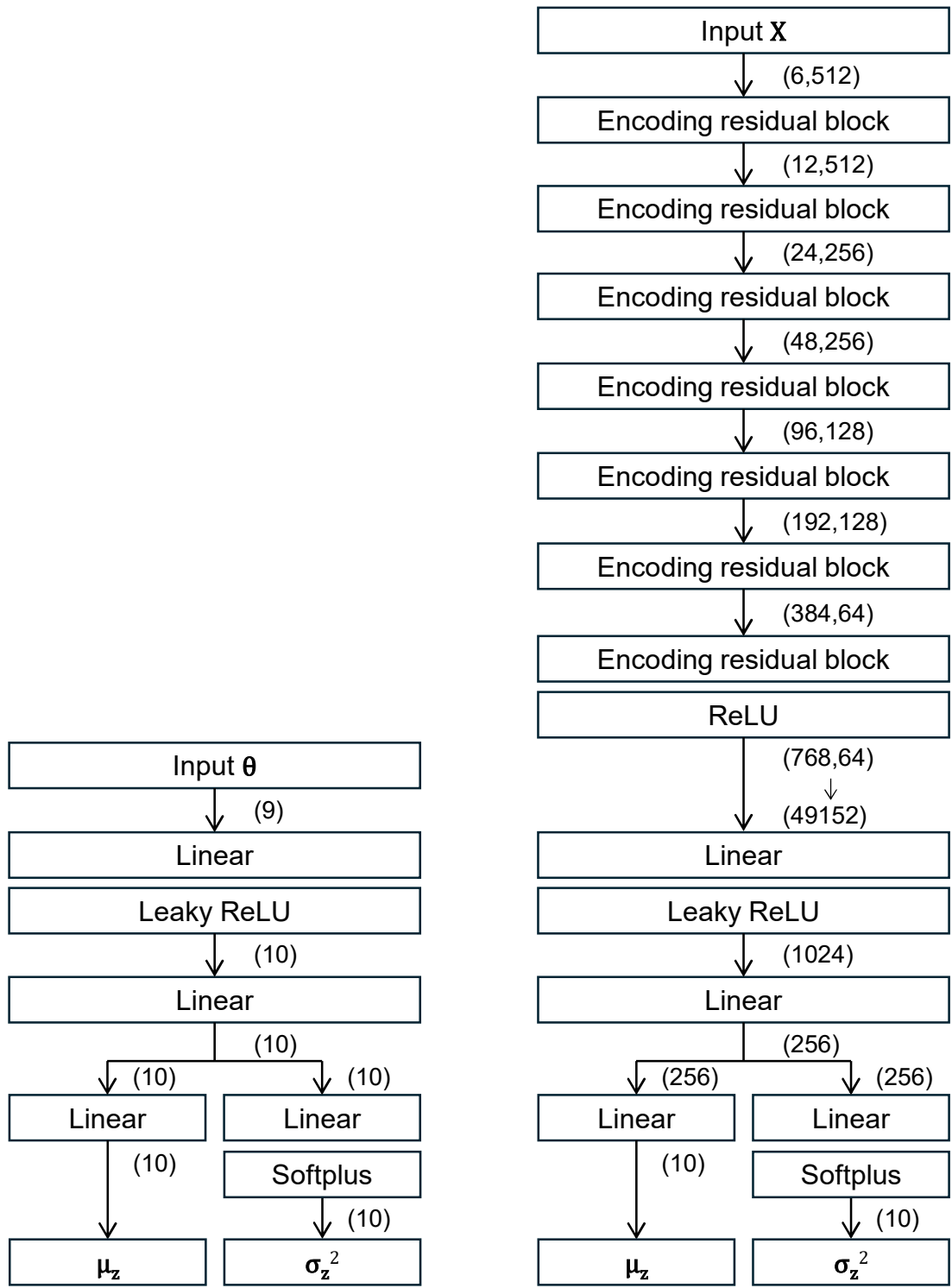

Fig. 16 Diagram of the surrogate unimodal encoders of JMVAE-kl used for lumped mass model

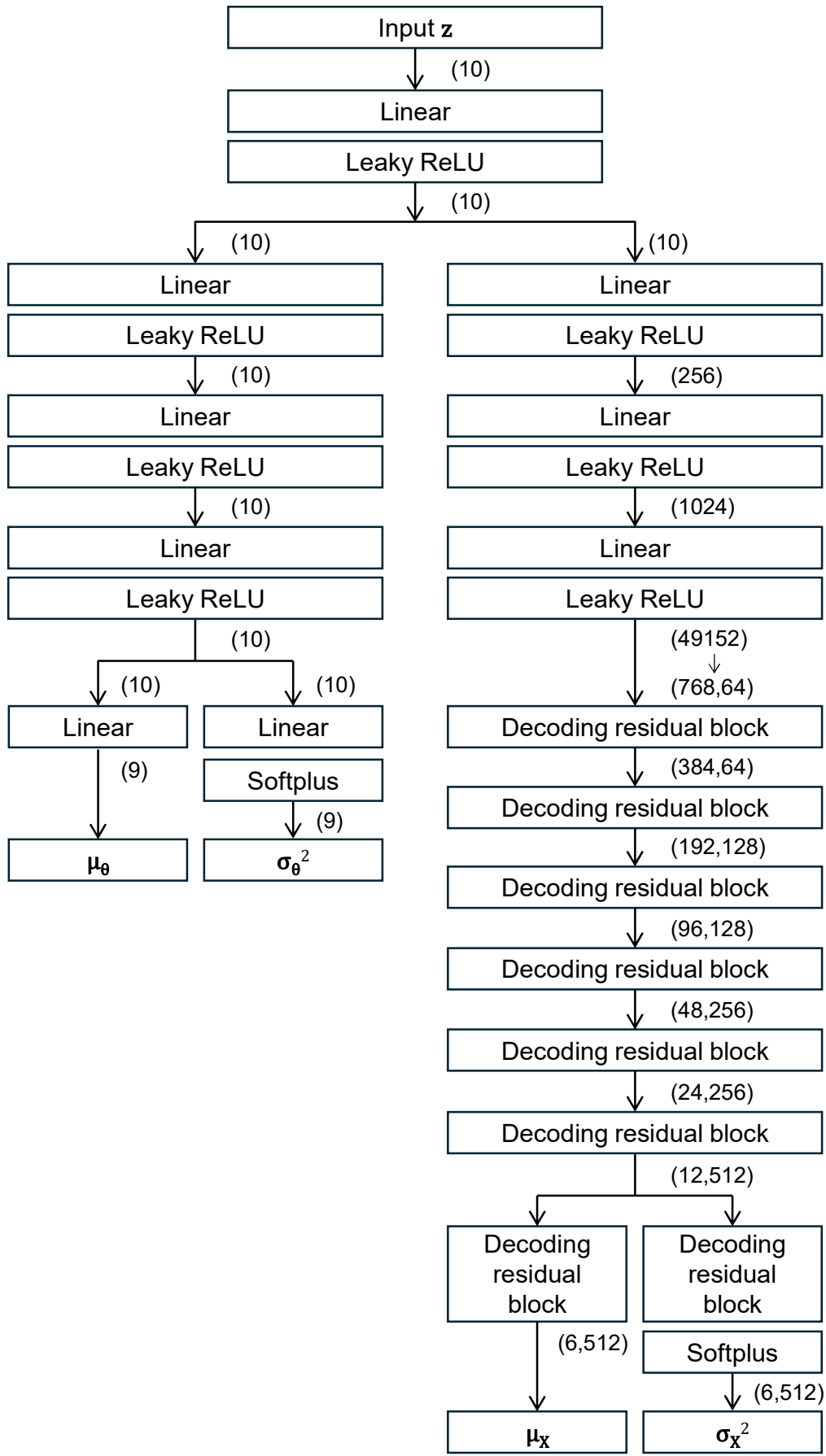

Fig. 17 Diagram of the decoder of JMVAE-kl used for lumped mass model

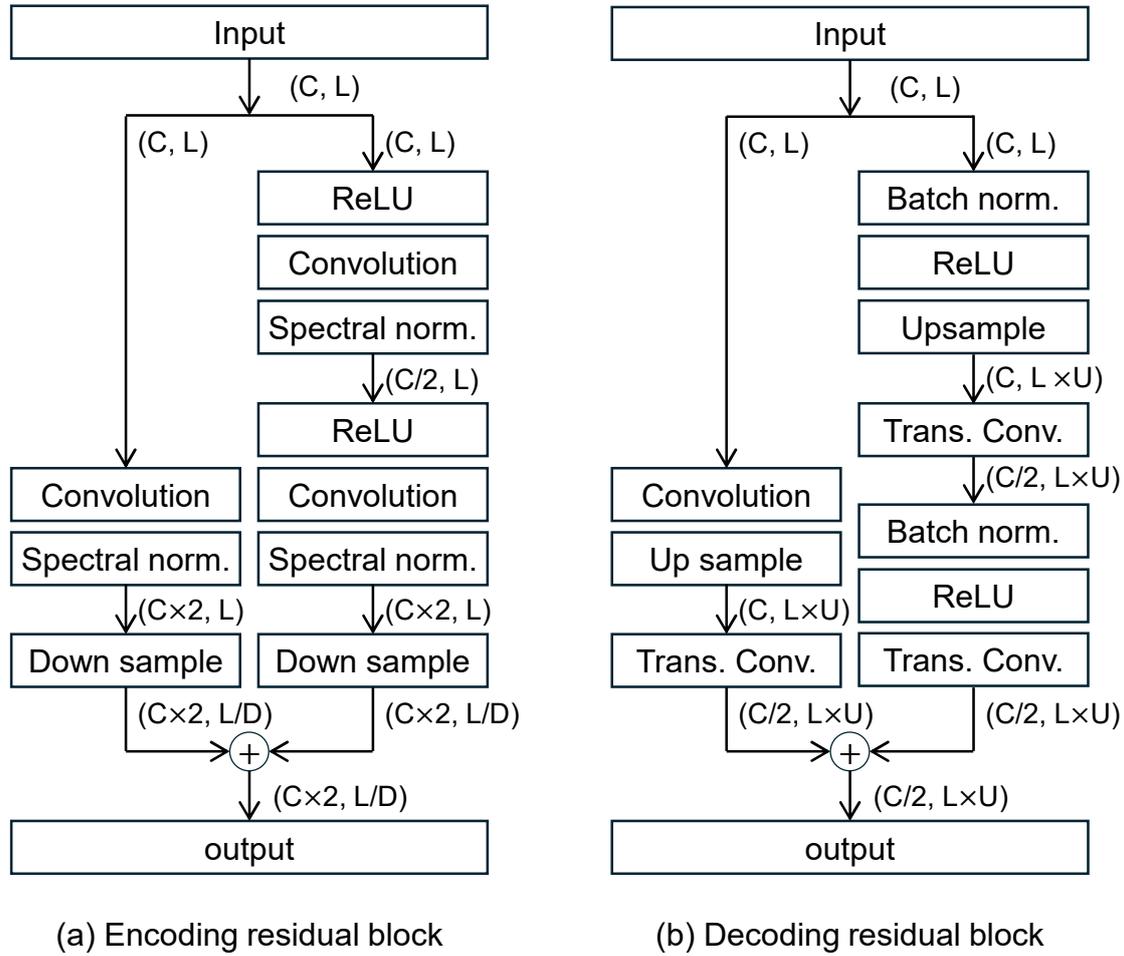

Fig. 18 Residual blocks in the VAE used for lumped mass model

4.3 Sampling from the posterior distribution of the model parameters

Sampling of the model parameters based on the posterior distribution was performed using the replica-exchange Monte Carlo method [38], [39], where a sample of each replica was generated based on the Metropolis-Hasting method. The replica exchange Monte Carlo method is an MCMC sampling method that is suitable for sampling multimodal and/or high-dimensional distributions. The exchange, according to the acceptance criteria, was repeated 1000 times for every 100 samples. The first 10000 samples were discarded as the burn-in samples. The samples were thinned by retaining every 30th sample. A total of 3000 samples were collected. The proposed method initializes each replica such that the sample in the latent variable space aligns closely with the observed data. For simplicity, a uniform distribution was assumed for the prior distribution $p(\boldsymbol{\theta})$, as in the example in Section 3.

    The training of the original VAE and sampling of the posterior distribution were completed, similar to the multimodal VAE case. For details on the VAE, please refer to [40].

    The cumulative distribution functions of the posterior distribution of the model parameters

by the multimodal and original VAE are shown in Fig. 19 and were compared with those of the ground-truth model. Both the multimodal and original VAE provided estimates consistent with the ground-truth model for initial stiffness $k_i$ ($i = 1, 2, 3$), crack displacement $d_c$, $\alpha_c$ and index $\gamma$, although a minor bias was observed in the case of $\alpha_c$. However, estimates of $d_y$, $\alpha_y$ and $\lambda$ by the multimodal VAE exhibited larger scatter than those by the original VAE. The scatter of the estimated model parameters primarily reflected the degree of uncertainty arising from the lack of information in the observations. Additionally, the scatter also reflected the model-related errors of surrogate unimodal model when approximating $p(\mathbf{z}|\boldsymbol{\theta})$. As discussed in Section 2, the calculation of the likelihood in case of the multimodal VAE was based on the already trained surrogate unimodal encoder for $\boldsymbol{\theta}$ to approximate $p(\mathbf{z}|\boldsymbol{\theta})$, which may affect accuracy. The larger scatters in cases of $d_y$, $\alpha_y$ and $\lambda$ were attributed to this error.

The computational time required to obtain the posterior distributions is listed in Table 4. The specifications of the Windows computer used are as follows: CPU: Intel i9-13900KF @ 3.00GHz, GPU: NVIDIA GeForce RTX 4090, and RAM: 128GB. In this case, the computational demands for estimating the posterior distribution were approximately 400 times lower, compared to the method utilizing the original VAE, although the reduction depended on the specific problem.

Table 4 Computational time required for estimating the posterior distribution (CPU: Intel i9-13900KF @ 3.00GHz, GPU: NVIDIA GeForce RTX 4090, RAM: 128GB)

| Proposed method using the multimodal VAE | Method utilizing the original VAE |
| --- | --- |
| 137.9 [s] | 55327.5 [s] |

Fig. 20 illustrates the response time histories of five randomly selected models with the model parameters estimated from the replica exchange Monte Carlo method using the original VAE. Both the acceleration time histories on the third floor (Fig. 20(a)) and the interstory drift of the first floor (Fig. 20(b)) exhibited amplitude and phase characteristics consistent with those of the ground-truth model shown in Fig. 14. Despite the varied model parameters, particularly for $d_y$, $\alpha_y$, $\gamma$, and $\lambda$, the consistency across models suggested that the observed data may not include information relevant to these parameters, that is, the parameters did not significantly influence the reproduction of the observation. Estimating the uncertainties of such parameters is crucial for predicting seismic responses to future ground motions. This represents a key advantage of the Bayesian model updating over deterministic approaches.

Fig. 21 illustrates the response time histories of the five models with model parameters estimated using the multimodal VAE. For the acceleration time histories on the third floor (Fig. 21(a)), both the amplitudes and phase characteristics aligned with those of the ground-truth model in Fig. 14(a). However, for the interstory drift of the first floor (Fig. 21(b)), slight discrepancies and scatter in the peak and residual drifts were observed, whereas the phase characteristics were consistent with those of the ground-truth model, as shown in Fig. 14(b). These discrepancies in the interstory drift were attributable to the scatter and bias in the estimated model parameters, as illustrated in Fig. 19. Although the original VAE model guarantees higher accuracy, a multimodal VAE is advantageous in certain scenarios when the structural analysis model is more dimensional and computational demands are higher. It should also be noted that in this example, a uniform distribution was assumed for the prior distribution $p(\boldsymbol{\theta})$ for simplification purposes. In practical applications, the assumption of an appropriate prior distribution of these parameters can improve the results.

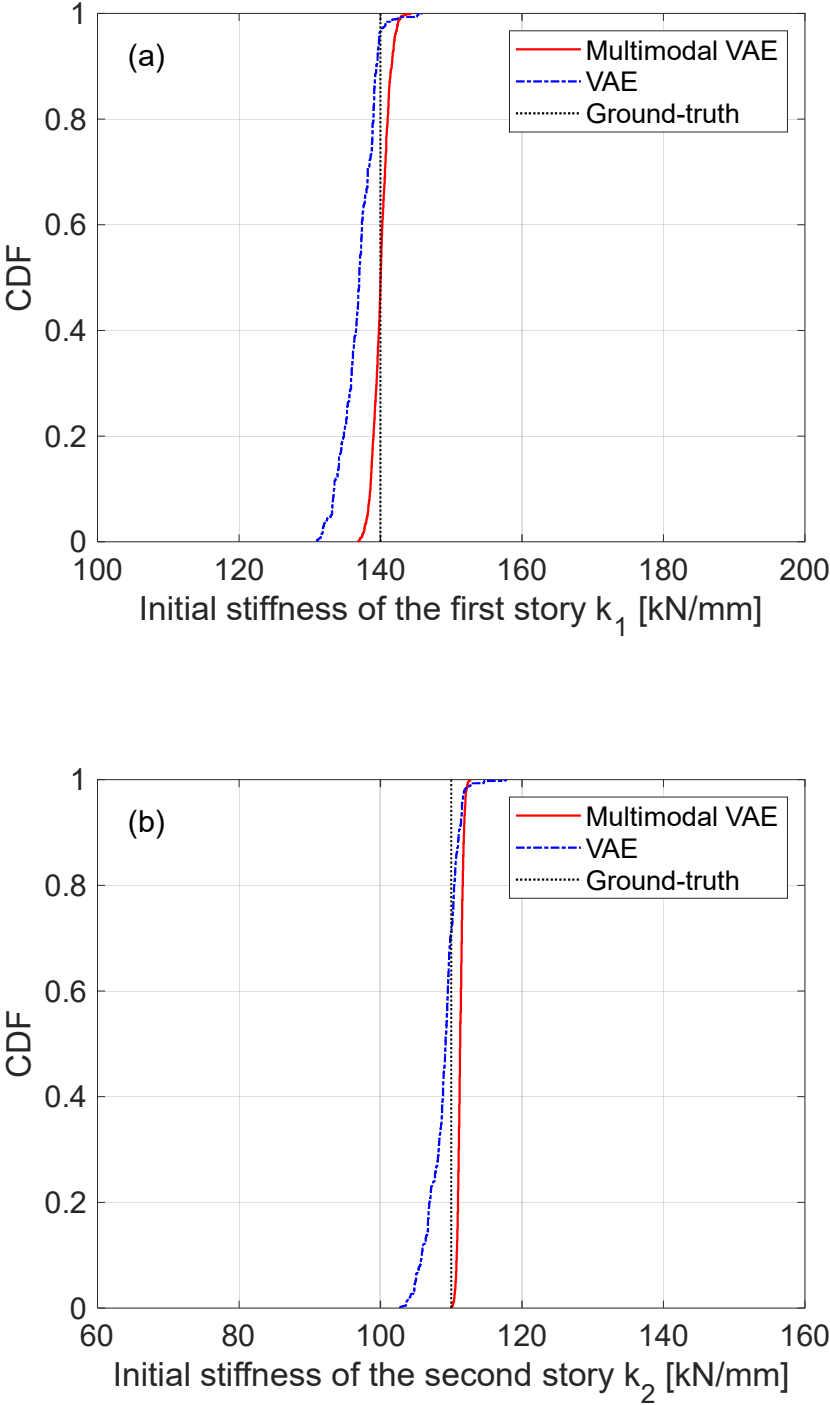

Fig. 19 Comparison between posterior distributions for model parameters for the three-degree-of-freedom model

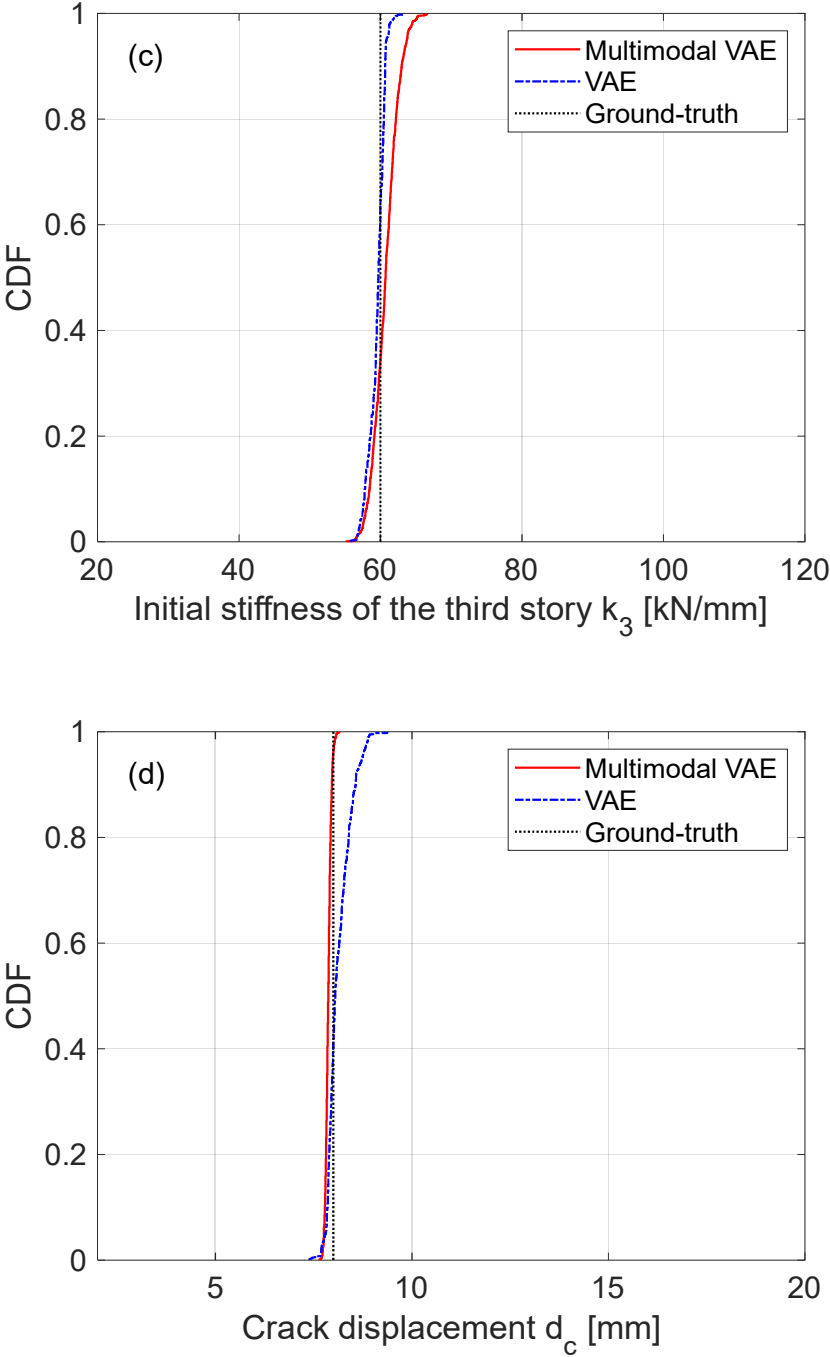

Fig. 19 Comparison between posterior distributions for model parameters for the three-degree-of-freedom model (cont.)

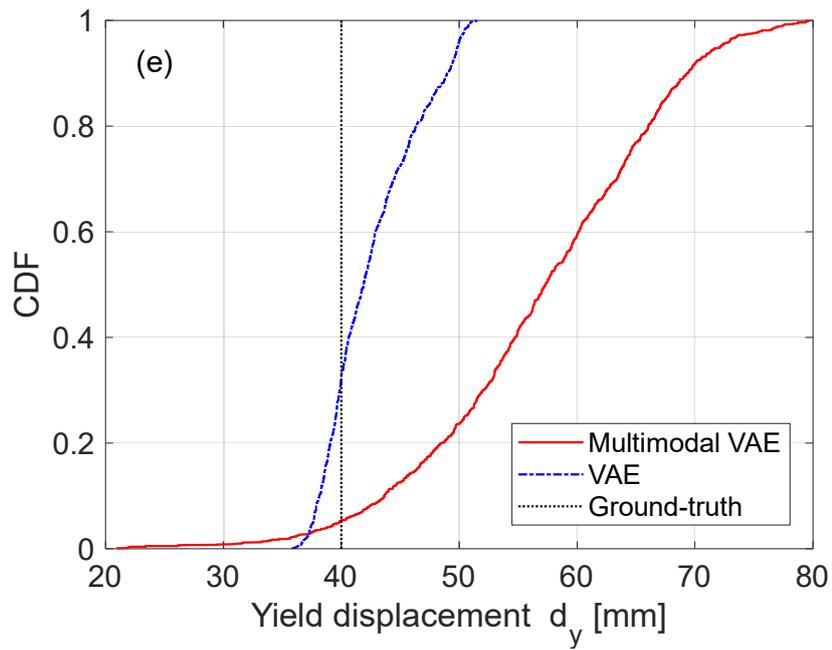

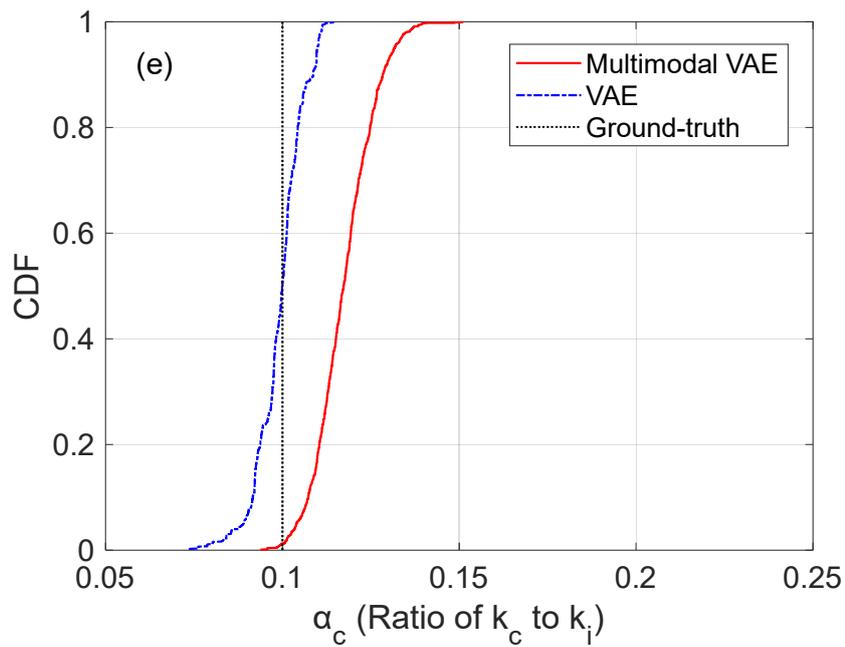

Fig. 19 Comparison between posterior distributions for model parameters for the three-degree-of-freedom model (cont.)

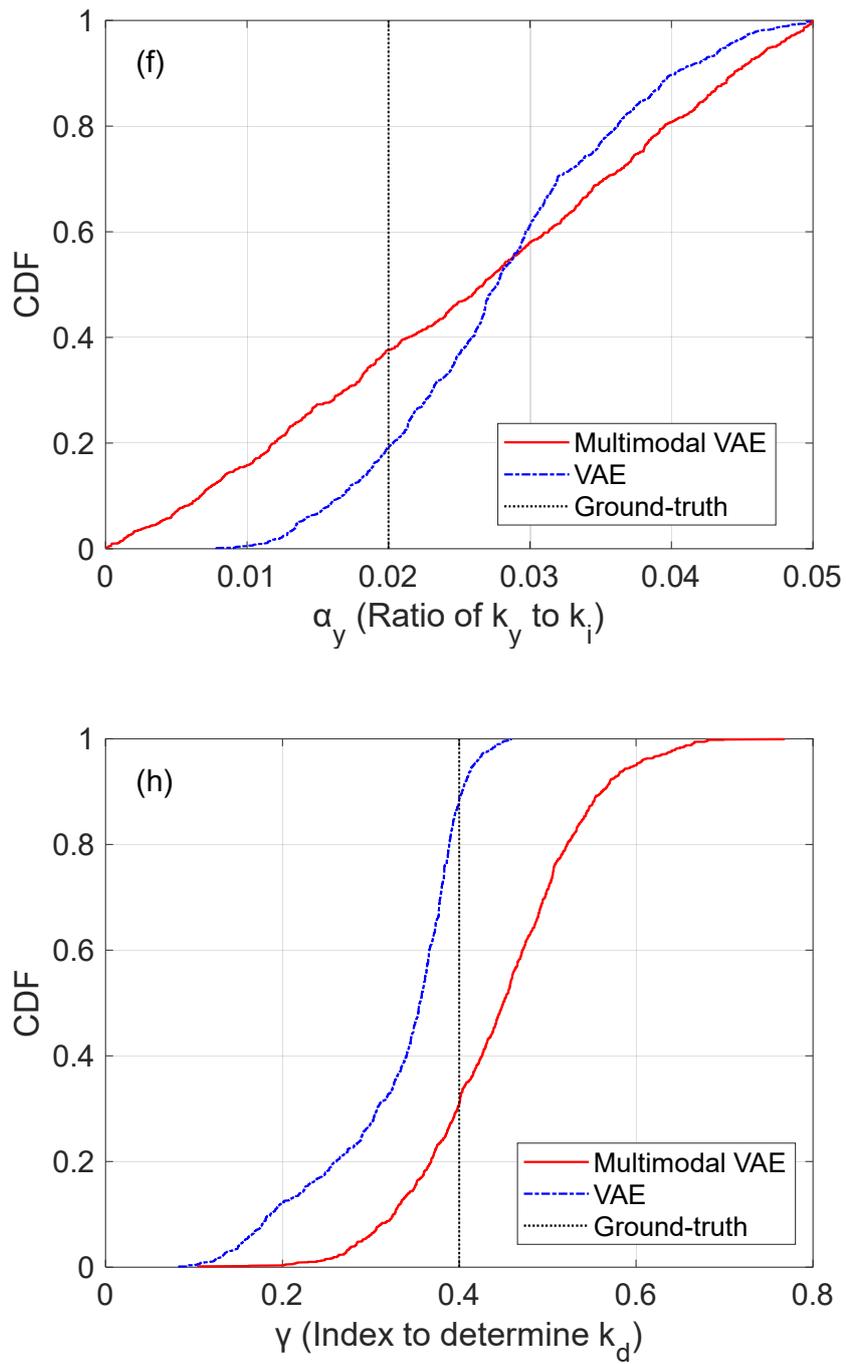

Fig. 19 Comparison between posterior distributions for model parameters for the three-degree-of-freedom model (cont.)

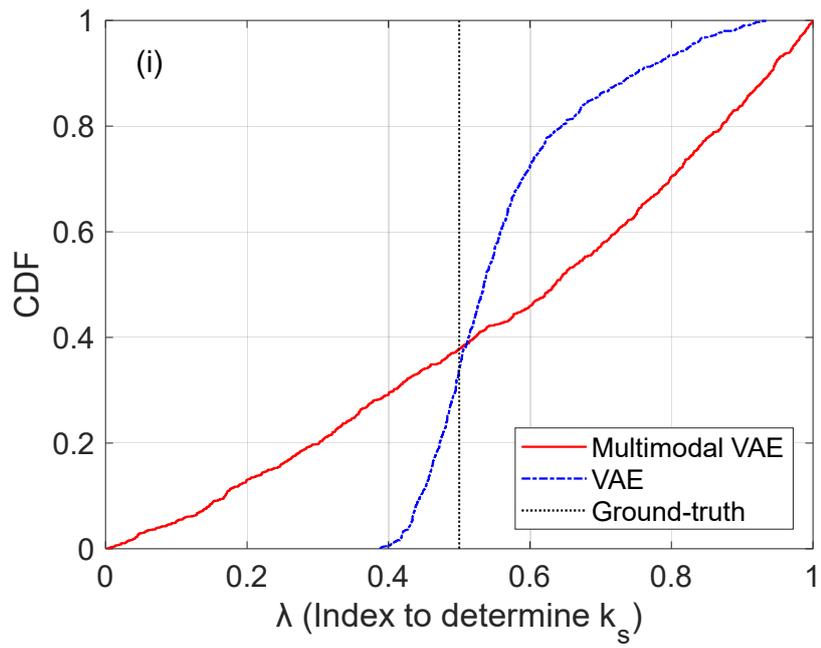

Fig. 19 Comparison between posterior distributions for model parameters for the three-degree-of-freedom model (cont.)

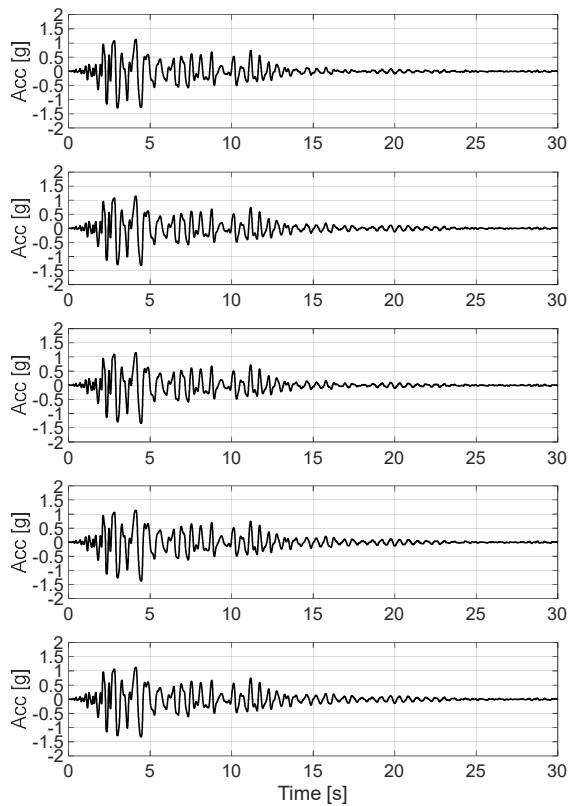

(a) Acceleration on third floor

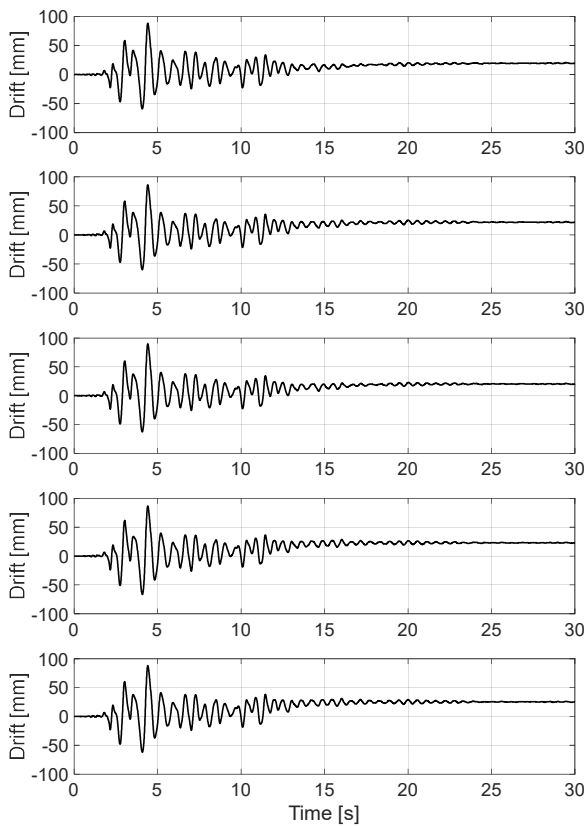

(b) Interstory drift on first floor

Fig. 20 Examples of response time history of the models with estimated model parameters in case of the original VAE

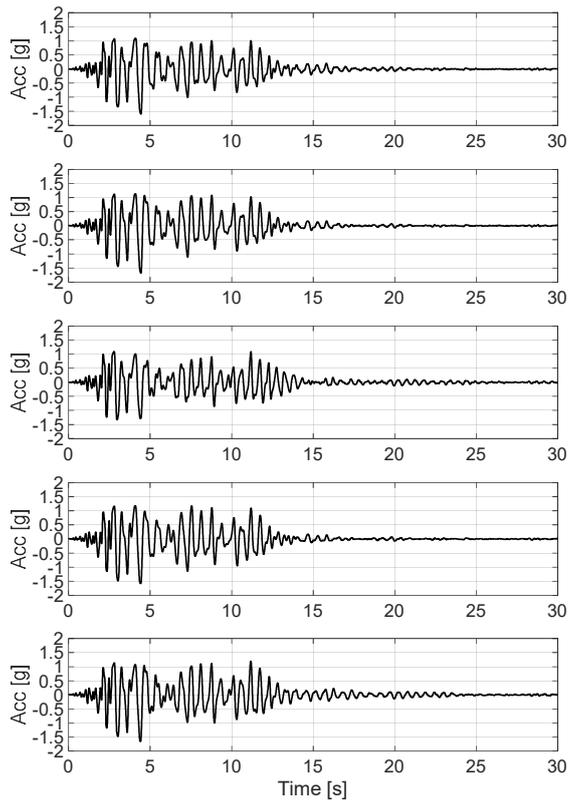

(a) Acceleration on third floor

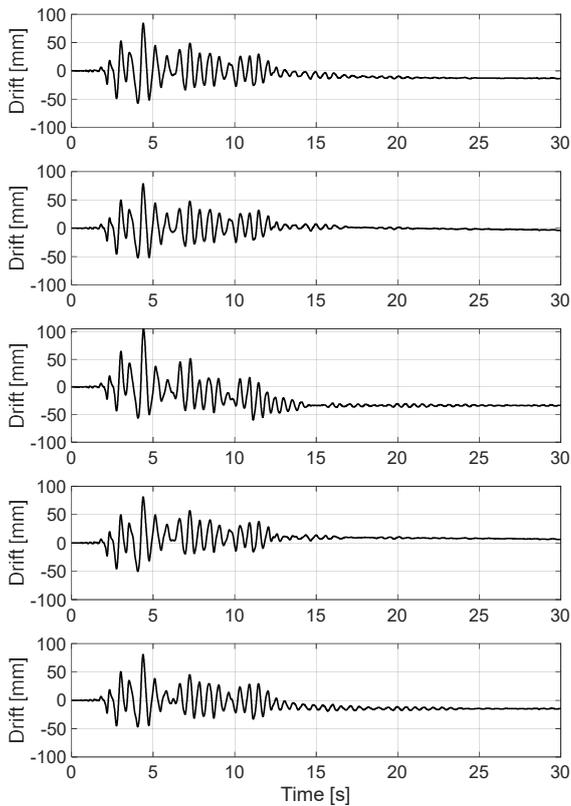

(b) Interstory drift on first floor

Fig. 21 Examples of response time history of the models with estimated model parameters in case of the multimodal VAE

5. Conclusions

A multimodal VAE that employs surrogate unimodal encoders of JMVAE-kl was proposed in this study as a versatile framework for Bayesian structural model updating. The proposed method is particularly suitable for simultaneous high-dimensional correlated observations used for structural model updating. A numerical example serves as a benchmark for the framework, demonstrating the updating of the model parameters of a single-story frame building structure equipped with acceleration and dynamic strain measurements. An additional numerical example demonstrates the Bayesian updating of nonlinear model parameters for a three-degrees-of-freedom lumped mass model. This example was used to compare the accuracy of the Bayesian model updating and the computational efficiency with those of the original VAE.

The degree of uncertainty in the updated model parameters depended on whether the observed data included parameter information. Furthermore, the estimated uncertainty in the model parameters reflected the model-related errors of the surrogate unimodal model in the multimodal VAE. The proposed Bayesian updating method utilizing the multimodal VAE successfully quantified the uncertainty in the estimated model parameters without a significant loss of accuracy compared with the original VAE. This suggests that model-related errors were not significant in the examples presented in this study. However, this conclusion is not universal and may vary depending on the arrangement and type of measurements, number of measurement points, and characteristics of the observed data, which will be explored in future studies. The multimodal VAE is particularly advantageous for Bayesian model updating when the structural analysis model is more high-dimensional and the computational demands are higher.

Data Availability
The data used to support the findings of this study are available from the corresponding author upon reasonable request.

Acknowledgments
A part of this study was supported by the MEXT Innovative Nuclear Research and Development Program [grant number JPMXD0222682433].